\documentclass{article}

\PassOptionsToPackage{numbers, compress}{natbib}
\usepackage[preprint]{neurips_2023}

\usepackage[utf8]{inputenc} % allow utf-8 input
\usepackage[T1]{fontenc}    % use 8-bit T1 fonts
\usepackage{hyperref}       % hyperlinks
\usepackage{url}            % simple URL typesetting
\usepackage{booktabs}       % professional-quality tables
\usepackage{amsfonts}       % blackboard math symbols
\usepackage{nicefrac}       % compact symbols for 1/2, etc.
\usepackage{microtype}      % microtypography
\usepackage{xcolor}         % colors
\usepackage{graphicx}
\usepackage{subcaption}
\usepackage{amsmath}

\title{MeSa: \underline{M}asked, G\underline{e}ometric, and \underline{S}upervised Pre-tr\underline{a}ining for Monocular Depth Estimation}
\newcommand{\METHOD}{MeSa}

\author{%
  Muhammad Osama Khan \\
  New York University\\
  \texttt{osama.khan@nyu.edu} \\
  \And
  Junbang Liang \\
  Amazon \\
  \texttt{junbanl@amazon.com} \\
  \And
  Chun-Kai Wang \\
  Amazon \\
  \texttt{ckwang@amazon.com} \\
  \And
  Shan Yang \\
  Amazon \\
  \texttt{ssyang@amazon.com} \\
  \And
  Yu Lou \\
  Amazon \\
  \texttt{ylou@amazon.com} \\
}

\begin{document}

\maketitle

\begin{abstract}
Pre-training has been an important ingredient in developing strong monocular depth estimation models in recent years. For instance, self-supervised learning (SSL) is particularly effective by alleviating the need for large datasets with dense ground-truth depth maps. However, despite these improvements, our study reveals that the later layers of the SOTA SSL method are actually suboptimal. By examining the layer-wise representations, we demonstrate significant changes in these later layers during fine-tuning, indicating the ineffectiveness of their pre-trained features for depth estimation. To address these limitations, we propose MeSa, a comprehensive framework that leverages the complementary strengths of masked, geometric, and supervised pre-training. Hence, MeSa benefits from not only general-purpose representations learnt via masked pre-training but also specialized depth-specific features acquired via geometric and supervised pre-training. Our CKA layer-wise analysis confirms that our pre-training strategy indeed produces improved representations for the later layers, overcoming the drawbacks of the SOTA SSL method. Furthermore, via experiments on the NYUv2 and IBims-1 datasets, we demonstrate that these enhanced representations translate to performance improvements in both the in-distribution and out-of-distribution settings. We also investigate the influence of the pre-training dataset and demonstrate the efficacy of pre-training on LSUN, which yields significantly better pre-trained representations. Overall, our approach surpasses the masked pre-training SSL method by a substantial margin of 17.1\% on the RMSE. Moreover, even without utilizing any recently proposed techniques, MeSa also outperforms the most recent methods and establishes a new state-of-the-art for monocular depth estimation on the challenging NYUv2 dataset.
\end{abstract}

\section{Introduction}
\label{sec:introduction}
Monocular depth estimation is an important computer vision problem, with applications ranging from self-driving cars to augmented reality and robotics. Initially, supervised learning methods~\citep{eigen2014depth,fu2018deep,laina2016deeper} were developed to tackle this problem, utilizing annotated depth data for training the models. However, collecting diverse real-world datasets with precise ground-truth depth is extremely challenging. Hence, self-supervised methods were developed that learn depth from stereo image pairs~\citep{garg2016unsupervised,godard2017unsupervised} or monocular videos~\citep{zhou2017unsupervised} without relying on depth annotations. Particularly, self-supervised monocular depth estimation from videos is especially appealing for real-world applications, as it only requires a single camera for data collection. These self-supervised methods typically employ view synthesis as the main supervision signal and are trained via the photometric loss~\citep{zhou2017unsupervised}.

In recent years, pre-training has emerged as an important factor in developing strong depth estimation models. For instance,~\citet{ranftl2020towards} noticed that randomly initialized models tend to be about 35\% worse than the same models pre-trained on ImageNet. Building on this,~\citet{xie2022revealing} recently obtained SOTA results by directly fine-tuning a SimMIM~\citep{xie2022simmim} pre-trained network for depth estimation. SimMIM~\citep{xie2022simmim}, a masked pre-training~\citep{he2022masked,xie2022simmim} method, learns self-supervised representations via the following pretext task: given a partially masked input image, the network reconstructs the masked portions of the image. Interestingly, this relatively simple pre-training task obtains SOTA performance on depth estimation without utilizing any geometric priors.

Despite these pre-training algorithms yielding significant improvements, we demonstrate that the last layers of the SOTA SSL method~\citep{xie2022revealing} are actually suboptimal. To investigate this, we compare the similarity of the pre-trained representation and fine-tuned representation for each layer (Figure~\ref{fig:sota_cka}). Intuitively, a higher similarity means that the pre-trained representation is well-suited for the downstream task and hence requires minimal updates during the fine-tuning process and vice versa~\citep{neyshabur2020being}. Our analysis reveals significant changes in these later layers during fine-tuning, indicating the ineffectiveness of the pre-trained features for depth estimation.

Based on the intuition that different pre-training strategies capture distinct types of features, we propose a comprehensive pre-training framework, MeSa, that addresses the aforementioned limitations. MeSa leverages the complementary strengths of three types of learning strategies: masked, geometric, and supervised pre-training. Via this synergy, our framework benefits from not only general-purpose representations learnt via masked pre-training but also specialized depth-specific features acquired via geometric and supervised pre-training.

Our layer-wise analysis confirms that our pre-training strategy indeed produces improved representations for the later layers, thereby successfully overcoming the drawbacks of the SOTA SSL method. Furthermore, via experiments on the NYUv2 and IBims-1 datasets, we demonstrate that these enhanced representations translate to performance improvements in both the in-distribution and out-of-distribution settings. Additionally, we investigate the impact of the pre-training dataset and demonstrate the superiority of pre-training on the LSUN dataset, resulting in significantly improved pre-trained representations. Moreover, via benefiting from the 3D projective geometry during pre-training, our method helps eliminate the artifacts observed in the SOTA SSL model that lacks geometric priors.

To sum up, our main contributions include:
\begin{itemize}
    \item A qualitative analysis of pre-training effectiveness based on layer-wise feature similarities that uncovers the insight that the later layers of the SOTA SSL method are not optimally pre-trained.
    \item A novel pre-training pipeline, MeSa, that effectively pre-trains the entire network by leveraging the complementary strengths of masked, geometric, and supervised pre-training, thereby benefiting from both general-purpose as well as specialized depth-specific features.
    \item Our pre-trained model surpasses the SOTA masked pre-training method by a substantial margin of 17.1\% on the RMSE. Moreover, even without utilizing any recently proposed techniques, it also outperforms the most recent methods and establishes a new state-of-the-art for monocular depth estimation on the challenging NYUv2 dataset.
\end{itemize}
\section{Related Work}
\label{sec:related_work}
\begin{figure}[t]
\centering
\includegraphics[width=0.98\textwidth]{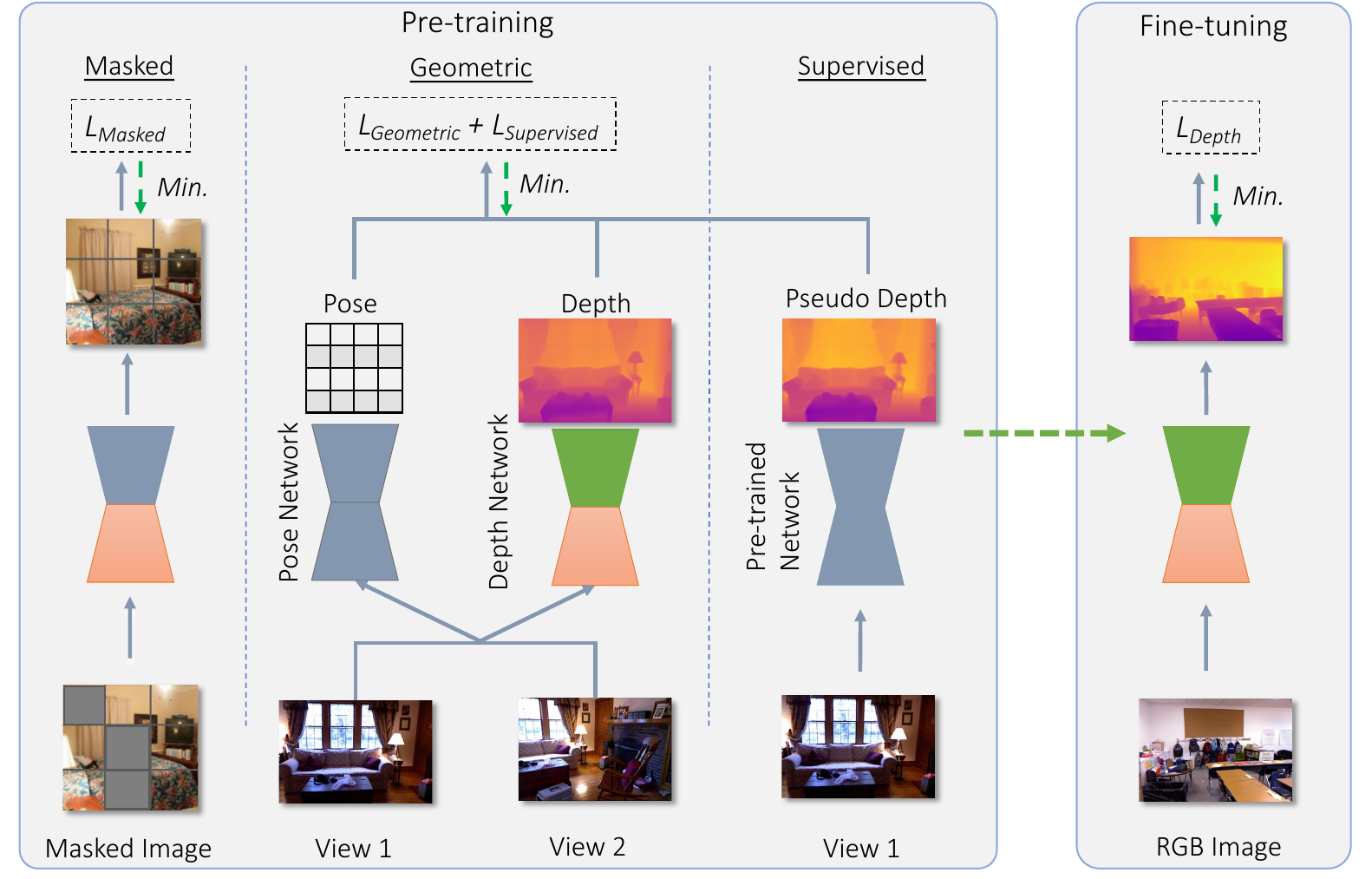}
\caption{Our proposed framework, MeSa, effectively leverages both general-purpose as well as depth-specific features via utilizing the complementary strengths of three pre-training strategies -- 1) Masked, 2) Geometric, and 3) Supervised pre-training. The pre-trained models are fine-tuned in a supervised manner on the downstream dataset.}
\label{fig:pipeline}
\end{figure}
\subsection{Monocular Depth Estimation}
Depth estimation was initially treated as a supervised learning problem~\citep{eigen2014depth,fu2018deep,laina2016deeper}, requiring ground-truth depth to train the networks. Since collecting ground-truth depth is a challenging problem in itself, some methods explored using synthetic data~\citep{mayer2018makes}. However, it is also not trivial to generate large varied synthetic datasets containing diverse scenes. A promising alternative is to train depth estimation networks using self-supervised learning, where view synthesis forms the main supervision signal and depth is synthesized as an intermediate step.~\citet{zhou2017unsupervised} developed one of the first monocular self-supervised algorithms, where they trained joint depth estimation and pose estimation networks using monocular videos. Since then, several methods~\citep{packnet,yin2018geonet,chen2019self,godard2017unsupervised,mahjourian2018unsupervised} have been proposed to improve various components. For instance, Monodepth2~\citep{monodepth2} proposed a simple model to elegantly handle occlusions as well as reduce visual artifacts.

SC-DepthV1~\citep{bian2021ijcv} proposed a geometry consistency loss in order to learn scale-consistent depth maps across the video. SC-DepthV2~\citep{bian2021tpami} built on top of this and proposed an auto-rectify network to alleviate training instabilities due to rotation of handheld cameras, thereby enabling better depth estimation on indoor scenes. Following this, SC-DepthV3~\citep{sun2022sc} used an external pre-trained depth estimation network to generate pseudo-depth, which was used to improve the depth estimation performance in dynamic scenes. Remarkably,~\citet{xie2022revealing} recently achieved SOTA performance by directly fine-tuning a masked pre-trained SimMIM~\citep{xie2022simmim} model for depth estimation without utilizing any geometric constraints used in previous methods. In this paper, we analyze the layer-wise representations of this SOTA model and leverage the 3D projective geometry in order to overcome its drawbacks, thereby designing a strong self-supervised learning algorithm for monocular depth estimation.
\subsection{Self-supervised Learning}
Self-supervised learning (SSL) is a powerful approach for learning representations from unlabeled data without requiring human annotation. SSL learns by leveraging various pretext tasks such as relative patch prediction~\citep{doersch2015context}, rotation prediction~\citep{gidaris2018rotation}, and colorization~\citep{zhang2016colorful}. A popular class of SSL methods is contrastive learning~\citep{wu2018memorybank,he2019moco,chen2020simclr,cao2020pic,grill2020byol,xie2021pixpro} which learns transformation invariant representations via maximizing the similarity of positive pairs and minimizing the similarity of negative pairs. Masked image modeling (MIM) has recently gained traction in vision~\citep{he2022masked,xie2022simmim} following the success of masked pre-training in NLP~\citep{devlin2018bert,liu2019roberta}. Although these MIM approaches have obtained SOTA results in several tasks, a comprehensive understanding of the pre-trained representations is still lacking, making the development of new SSL methods challenging. Recently, a few methods~\citep{xie2022revealing,pasad2021layer,pasad2023comparative} have attempted to understand these pre-trained presentations. Our work follows in this direction and uses the insights from our analysis to design a better self-supervised learning algorithm that effectively pre-trains the entire network for depth estimation.

\section{Method}
\label{sec:method}
As illustrated in Figure~\ref{fig:pipeline}, our framework consists of three pre-training strategies -- 1) Masked, 2) Geometric, and 3) Supervised pre-training. By synergistically integrating these three strategies, MeSa facilitates effective representation learning that benefits from both general-purpose representations (via masked pre-training) as well as depth-specific features (via geometric and supervised pre-training). Hence, MeSa effectively pre-trains the entire network including the later layers, thereby addressing the limitations of the SOTA SSL method. In the following subsections, we briefly introduce masked, geometric, and supervised pre-training in Sections~\ref{sec:mp},~\ref{sec:gp}, and~\ref{sec:sp} respectively. This is then followed by the fine-tuning and implementation details in Section~\ref{sec:implementation_details}. Lastly, Section~\ref{sec:layer_wise_analysis} outlines the techniques used for the layer-wise analysis to qualitatively understand the representations learnt via the three pre-training strategies.

\subsection{Masked Pre-training}
\label{sec:mp}
Masked pre-training learns representations by masking a region of the input image and reconstructing the masked region based on the partially masked input. In this work, we utilize the SimMIM framework proposed by~\citet{xie2022simmim} for masked pre-training. However, other pre-training methods could be used as well since our framework is agnostic to the exact choice of pre-training method. We use a simple masking strategy and randomly mask out some patches of the input image. The decoder consists of a single linear layer to predict the pixels of the masked regions. Finally, the network is trained using a plain $\ell_1$ loss applied to the masked regions:
\begin{equation}
L=\frac{1}{|\mathbf{x}_M|}\left\|\mathbf{x}_M-g(f(\hat{\mathbf{x}}))_M\right\|_1
\end{equation}
where $f$ is the encoder, $g$ is the decoder, $\mathbf{x}$ is the raw image, and $\hat{\mathbf{x}}$ is the partially masked image input to the encoder. The subscript $M$ denotes the masked subset of pixels whereas $|\cdot|$ denotes the number of pixels. After training, the decoder $g$ is usually discarded and the encoder $f$ is used for downstream tasks.

\subsection{Geometric Pre-training}
\label{sec:gp}
Geometric pre-training learns representations via utilizing the 3D projective geometry. In this work, we use the SC-DepthV2~\citep{bian2021tpami} framework for geometric pre-training, where depth and pose networks are trained jointly on monocular videos. Given consecutive image pairs $I_a$, $I_b$ from a video, we first generate the predicted depths $D_a$, $D_b$ of the two views and the relative camera pose $P_{ab}$ between them:
\begin{equation}
D_a=g'(f(I_a)),
   \quad
D_b=g'(f(I_b)),
   \quad
P_{ab}=h(I_a,I_b)
\end{equation}
where $f$ and $g'$ are the depth estimation encoder and decoder respectively whereas $h$ is the pose estimation network.

View synthesis forms the main supervision signal (i.e., given one view, the task is to predict a view of the same scene from a different viewpoint). Given the predicted depth $D_a$, relative camera pose $P_{ab}$, and the source view $I_b$, differentiable bilinear interpolation~\citep{jaderberg2015stn} is used to generate a prediction of the reference view $I_{a'}$. Hence, depth is actually synthesized as an intermediate step and photometric loss between the generated $I_{a'}$ and actual $I_{a}$ images is used to train the network:
\begin{equation}
L_P=\frac{1}{|\mathcal{V}|} \sum_{p \in \mathcal{V}}\left(\lambda\left\|I_a(p)-I_a^{\prime}(p)\right\|_1+(1-\lambda) \frac{1-\operatorname{SSIM}_{a a^{\prime}}(p)}{2}\right)
\end{equation}
where $\mathcal{V}$ is the set of valid points projected from $I_{a}$ to $I_{b}$ and $\operatorname{SSIM}_{a a^{\prime}}$ is the similarity between $I_{a}$ and $I_{a'}$ computed via the SSIM function~\citep{wang2004image}.

Moreover, the geometry consistency loss is used to ensure that the two depth maps ($D_a$ and $D_b$) are consistent in terms of 3D geometry:
\begin{equation}
L_G=\frac{1}{|\mathcal{V}|} \sum_{p \in \mathcal{V}} D_{\mathrm{diff}}(p)
\end{equation}
where $D_{\mathrm{diff}}$ is the depth inconsistency map between $D_{a}$ and $D_{b}$. For further details, please refer to SC-Depth~\citep{bian2021ijcv}.
\subsection{Supervised Pre-training}
\label{sec:sp}
Supervised pre-training learns representations via using off-the-shelf supervised pre-trained networks for depth estimation. In this work, we use the SC-DepthV3~\citep{sun2022sc} framework for supervised pre-training, which uses the LeReS~\citep{yin2021learning} pre-trained network to generate pseudo-depth. This pseudo-depth is used to improve performance on dynamic objects via the confident depth ranking loss ($L_{CDR}$) and on object boundaries via the egde-aware relative normal loss ($L_{ERN}$). Hence, the overall loss for joint training via geometric and supervised pre-training is given by:
\begin{equation}
L=\alpha L_P^M+\beta L_G+\gamma L_N+\delta L_{\mathrm{CDR}}+\epsilon L_{\mathrm{ERN}}
\end{equation}
where $L_{P}^{M}$ is the weighted photometric loss, $L_{G}$ is the geometry consistency loss, $L_{N}$ is the normal matching loss, $L_{CDR}$ is the confident depth ranking loss, and $L_{ERN}$ is the edge-aware relative normal loss.
\subsection{Implementation Details}
\label{sec:implementation_details}
We pre-train via the three learning strategies sequentially in order to avoid training instability associated with concurrent training~\citep{guo2022discriminative}. Moreover, this also allows us to significantly reduce the pre-training time since we can leverage existing pre-trained models, thereby eliminating the need for training the entire pipeline from scratch whenever a new pre-training method is added.

In the masked pre-training phase, we employ a Swin-v2-L network as the encoder $f$ and a single linear layer as the decoder $g$. For the subsequent geometric and supervised pre-training stages, we keep the pre-trained encoder $f$ and replace the decoder with a DispNet~\citep{zhou2017unsupervised} $g'$. Additionally, we utilize a ResNet18~\citep{he2016deep} backbone in the pose network $h$ to compute the relative camera pose between the concatenated input images.

We use the LSUN~\citep{yu2015lsun} dataset for masked pre-training whereas we utilize the training split of the NYUv2~\citep{silberman2012indoor} dataset for geometric pre-training. For supervised pre-training, we leverage the pre-trained network provided by LeReS~\citep{yin2021learning} to generate the pseudo-depth. We follow the training details outlined in the official methods (SimMIM~\citep{xie2022simmim} and SC-Depth~\citep{sun2022sc}), with the exception of using an image size of 480$\times$480 instead of 256$\times$320 for geometric and supervised pre-training on NYUv2.

To evaluate all the pre-trained methods, we follow the SOTA SSL method~\citep{xie2022revealing} and fine-tune them on the training split of the NYUv2 dataset~\citep{silberman2012indoor}. We fine-tune for 75 epochs on 8 A100 GPUs, using a polynomial learning rate schedule with a 0.9 factor and a min lr of $10^{-5}$ and a max lr of $10^{-4}$.
\subsection{Layer-wise Analysis}
\label{sec:layer_wise_analysis}
\begin{figure}[t]
\begin{minipage}{.5\linewidth}
\centering
    \includegraphics[width=0.75\textwidth]{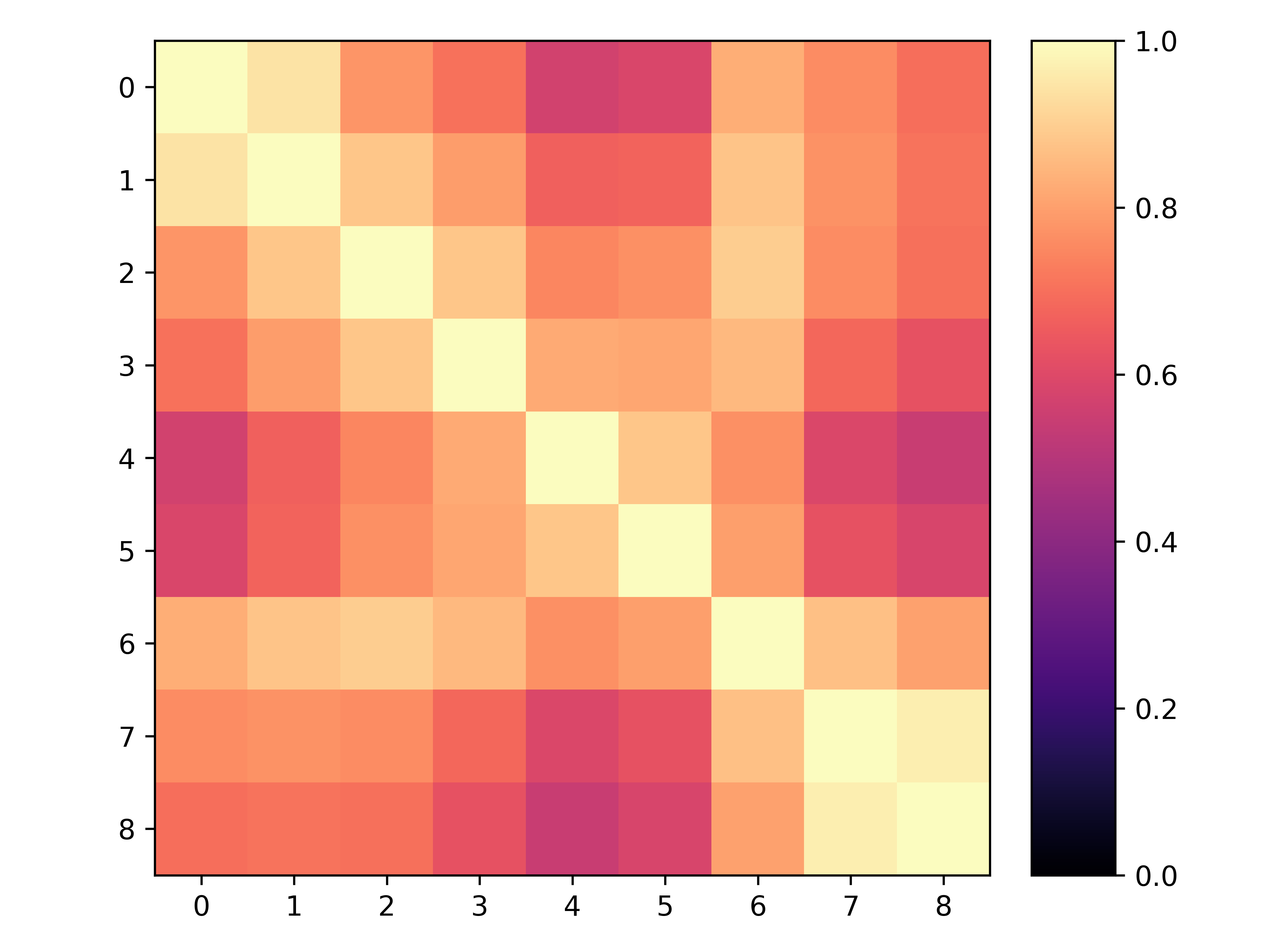}
  \vspace{10pt}
  \begin{picture}(0,0)
    \put(-150,15){\rotatebox{90}{{\footnotesize \texttt{Pre-trained Layers}}}}
    \put(-125,-8){{\footnotesize \texttt{Pre-trained Layers}}}
  \end{picture}
\end{minipage}\hfill
\begin{minipage}{.5\linewidth}
\centering
    \includegraphics[width=0.75\textwidth]{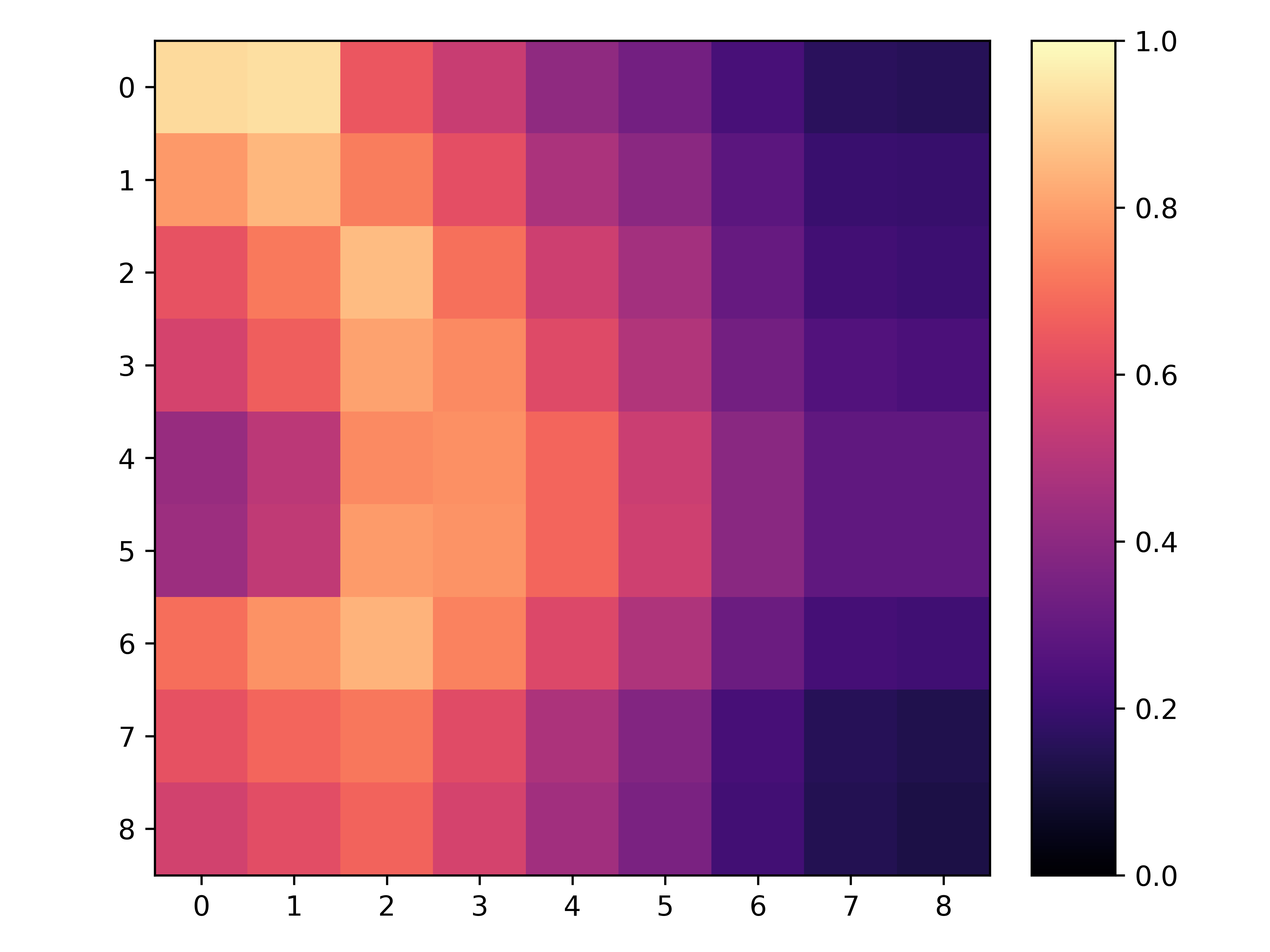}
  \vspace{10pt}
  \begin{picture}(0,0)
    \put(-150,15){\rotatebox{90}{{\footnotesize \texttt{Pre-trained Layers}}}}
    \put(-125,-8){{\footnotesize \texttt{Fine-tuned Layers}}}
  \end{picture}
\end{minipage}
\caption{Layer-wise analysis of the SOTA masked pre-trained model. \textbf{Left}: We discover a distinctive U-shaped pattern (top row) in representations of the pre-trained model as we delve deeper into the network. \textbf{Right}: We observe that the later layers (5-8) undergo significant changes during fine-tuning, as depicted by the lower similarities between pre-trained and fine-tuned features along the diagonal line. These results indicate that the pre-trained representations of the later layers are not effectively utilized for the downstream depth estimation task.}
\label{fig:sota_cka}
\end{figure}
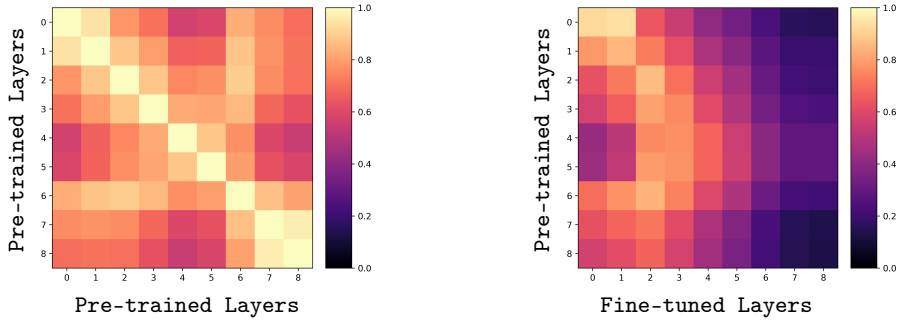
Following previous work~\citep{xie2022revealing,neyshabur2020being}, we use CKA (centered kernel alignment)~\citep{kornblith2019similarity} for our analysis, which is a metric that allows us to compare the representations of various layers within a network. CKA similarity values range from 0 to 1, with increasing values meaning that the two representations are more similar. We compare the layer-wise representations of nine layers within the Swin-v2-L architecture. For convenience, we refer to them as layers 0-8, with the exact layers detailed in the supplementary material.

We perform two types of analyses to evaluate the efficacy of the pre-trained layer-wise representations. Firstly, we compare the representations of layers 1-8 in the network to layer 0 (Figure~\ref{fig:sota_cka} left) in order to study the changes in representations as we delve deeper into the network. Secondly, for each layer, we compare the similarity of its pre-trained representation to its fine-tuned representation (Figure~\ref{fig:sota_cka} right) in order to understand how each layer evolves during the fine-tuning process. Intuitively, a higher similarity means that the pre-trained representation is well-suited for the downstream task and hence requires minimal updates during the fine-tuning process and vice versa~\citep{neyshabur2020being}.
\section{Experiments}
\label{sec:experiments}
Firstly, we analyze the layer-wise representations of the SOTA SSL model in order to illustrate the suboptimal pre-trained representations of the later layers (Section~\ref{sec:sota_cka}). To address this shortcoming, we propose a novel pre-training pipeline, MeSa, that not only achieves improved quantitative performance in both in-distribution (Section~\ref{sec:magis}) and out-of-distribution (Section~\ref{sec:ood}) settings but also learns better layer-wise representations (Section~\ref{sec:magis_cka}). Lastly, we investigate the impact of different pre-training datasets on the layer-wise features (Section~\ref{sec:lsun_pretraining}) and conclude with a comprehensive comparison against the SOTA depth estimation models (Section~\ref{sec:sota}).
\subsection{Layer-wise Analysis of the SOTA SSL Model}
\label{sec:sota_cka}
\begin{table}[b]
\caption{Geometric and supervised pre-training have complementary benefits to masked pre-training and result in significant improvements. MP: masked pre-training, GP: geometric pre-training, SP: supervised pre-training.}
\label{table:nyu}
\centering
\begin{tabular}{lcccccc}
\toprule Pre-training Strategy & RMSE $\downarrow$ & $\delta_1 \uparrow$ & $\delta_2 \uparrow$ & $\delta_3 \uparrow$ & $\mathrm{REL} \downarrow$ & $\log 10 \downarrow$ \\
\midrule
MP~\citep{xie2022revealing} & 0.287 & 0.949 & 0.994 & 0.999 & 0.083 & 0.035 \\
MP + GP & 0.269 & 0.951 & 0.993 & 0.998 & 0.076 & 0.033 \\
MP + GP + SP (MeSa) & \textbf{0.265} & \textbf{0.954} & \textbf{0.995} & \textbf{0.999} & \textbf{0.074} & \textbf{0.032} \\\midrule
Relative Improvement (\%) & 7.64 & 0.54 & 0.06 & 0 & 10.2 & 8.05 \\
\bottomrule
\end{tabular}
\end{table}
Although the SOTA SSL model yields excellent performance, it is unclear which parts of the pre-trained model contribute to this improvement. In order to understand this, we analyze the layer-wise representations of this masked pre-trained model using the techniques discussed in Section~\ref{sec:layer_wise_analysis}.

Based on the analysis in Figure~\ref{fig:sota_cka} (left), we discover a U-shaped pattern in the representation space, where the representations initially start getting farther apart but then become more similar to the initial layers towards the end of the network. The top row of Figure~\ref{fig:sota_cka} (left) compares the representation of layer 0 against the representations of layers 0-8. The similarity of layer 0’s representation with itself is 1 as expected. Thereafter, for layers 1-4, the representations begin to diverge from layer 0's representation, as indicated by the decreasing similarities (darker colors). However, interestingly, past layer 4, the similarity increases (lighter colors) suggesting that the representations are becoming more similar to the layer 0 representation. This U-shaped pattern is likely a result of the pre-training objective, which is to reconstruct masked portions of the image. Hence, the later layers end up having similar representations to the first few layers. A similar pattern has also been observed in~\citep{pasad2021layer,pasad2023comparative} in the context of speech processing.

Whereas this U-shaped pattern is useful for solving the self-supervised pretext task, it is not obvious if it aids in learning effective representations for downstream tasks. To investigate this, we compare the similarity of the pre-trained and fine-tuned representations of each layer in Figure~\ref{fig:sota_cka} (right). From the diagonal line (top-left to bottom-right), we observe that the similarities are quite small for the later layers (5-8). Lower similarities imply that these layers undergo significant changes during fine-tuning, indicating that they are not effective for the downstream depth estimation task~\citep{neyshabur2020being}.
\subsection{MeSa}
\label{sec:magis}
To address the aforementioned drawbacks, we propose a novel pre-training pipeline, MeSa, that utilizes the complementary benefits of masked, geometric, and supervised pre-training.

Table~\ref{table:nyu} shows the depth estimation results of the three pre-training strategies on the NYUv2 dataset. The results demonstrate significant performance improvements achieved by all three learning strategies, highlighting their complementary strengths. Overall, our approach surpasses the SOTA SSL model, which relies solely on masked pre-training, by 10.2\% on absolute relative error (0.083$\rightarrow$0.074) and 7.64\% on RMSE (0.287$\rightarrow$0.265).
\begin{figure}[t]
  \centering
  \renewcommand{\arraystretch}{0.0} % decrease vertical spacing
  \begin{tabular}{c}
    \begin{subfigure}{0.89\textwidth}
      \centering
      \includegraphics[width=\textwidth]{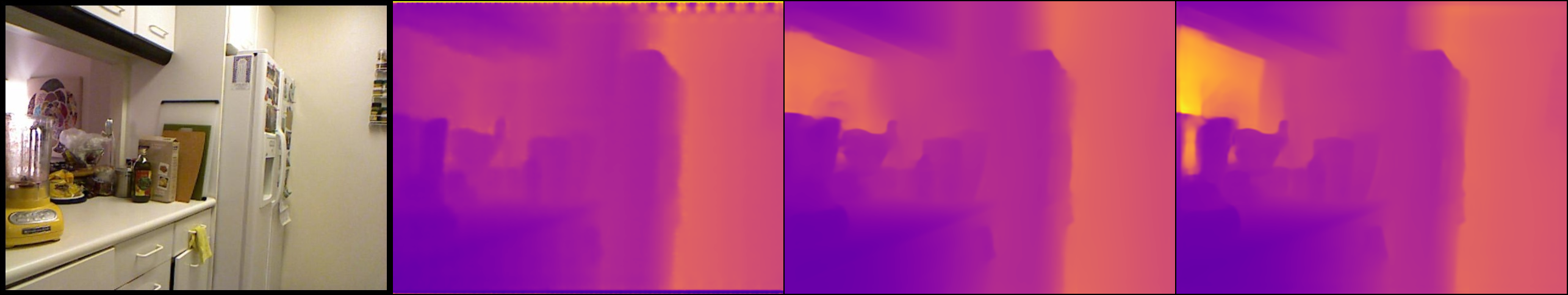}
    \end{subfigure} \\
    \begin{subfigure}{0.89\textwidth}
      \centering
      \includegraphics[width=\textwidth]{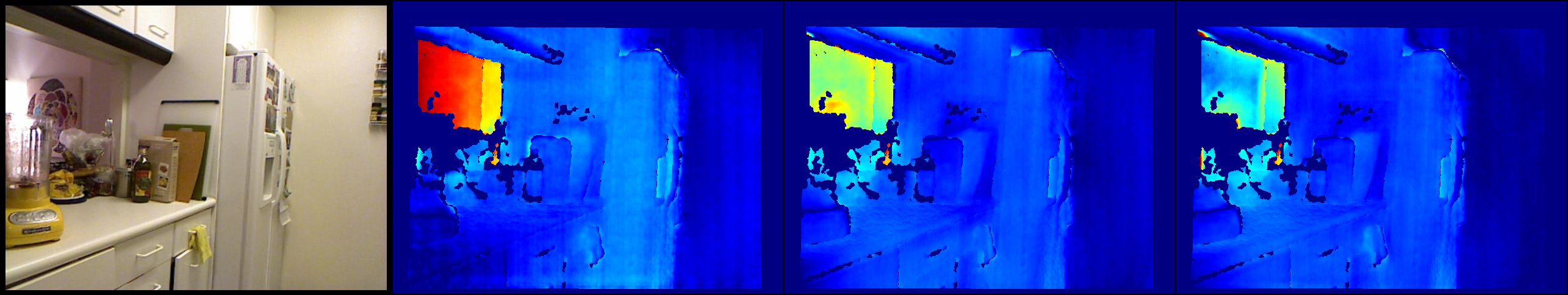}
    \end{subfigure} \\
  \end{tabular}
  \caption{Qualitative comparisons of the three pre-training strategies on NYUv2. Left$\rightarrow$right: Image, MP, MP+GP, MP+GP+SP (MeSa). \textbf{Top}: predicted depth maps, \textbf{bottom}: error maps (blue$\rightarrow$lower error; red$\rightarrow$higher error).}
  \label{fig:nyu_vis}
\end{figure}
Figure~\ref{fig:nyu_vis} presents qualitative results of the three pre-training strategies on the NYUv2 dataset, demonstrating significant improvements through the integration of geometric and supervised pre-training. When relying solely on masked pre-training, as in the SOTA SSL method, artifacts appear at the top and bottom of the predicted depth maps due to the absence of ground-truth depth (during fine-tuning) in these regions. By utilizing geometric pre-training, our method effectively eliminates these artifacts since the photometric loss utilizes 3D projective geometry and learns accurate depth across the entire image without relying on ground-truth. Moreover, geometric and supervised pre-training also help enhance the sharpness of the predicted depth maps.
\subsection{Out-of-distribution Performance}
\label{sec:ood}
In this section, we evaluate the performance of the three pre-training strategies in the out-of-distribution (OOD) setting. To this end, we evaluate the models trained on the NYUv2 dataset directly on the IBims-1 dataset without any additional fine-tuning.
\begin{table}[b]
\caption{In addition to improving the overall depth estimation accuracy, geometric and supervised pre-training also enhance the accuracy of depth boundaries and planar regions in the OOD setting. MP: masked pre-training, GP: geometric pre-training, SP: supervised pre-training.}
\label{table:ibims}
\centering
\begin{tabular}{lccccc}
\toprule Pre-training Strategy & $\varepsilon_{\mathrm{DBE}}^{\text {acc }} \downarrow$ & $\varepsilon_{\mathrm{DBE}}^{\text {comp }} \downarrow$ & $\varepsilon_{\mathrm{PE}}^{\text {plan }} \downarrow$ & $\varepsilon_{\mathrm{PE}}^{\text {orie }} \downarrow$ & AbsRel $\downarrow$ \\
\midrule
MP~\citep{xie2022revealing} & 2.40 & 30.05 & 3.26 & 8.16 & 0.089 \\
MP + GP & 2.25 & 24.69 & 2.46 & \textbf{6.29} & \textbf{0.082} \\
MP + GP + SP (MeSa) & \textbf{2.16} & \textbf{19.31} & \textbf{2.17} & 6.44 & 0.083 \\\midrule
Relative Improvement (\%) & 10.3 & 35.8 & 33.4 & 22.9 & 8.4 \\
\bottomrule
\end{tabular}
\end{table}

Table~\ref{table:ibims} presents the OOD results on the IBims-1 dataset. In addition to the overall depth estimation accuracy (measured via AbsRel), it is also important to improve the accuracy of depth boundaries (measured via $\varepsilon_{\mathrm{DBE}}^{\text {acc }}$ and $\varepsilon_{\mathrm{DBE}}^{\text {comp }}$) as well as planar regions (measured via $\varepsilon_{\mathrm{PE}}^{\text {plan }}$ and $\varepsilon_{\mathrm{PE}}^{\text {orie }}$), both of which are critical for many real-world applications. Our results highlight that we surpass the SOTA masked pre-training method on all five metrics in the OOD setting. Notably, we achieve substantial improvements of 35.8\% (30.05$\rightarrow$19.31) and 33.4\% (3.26$\rightarrow$2.17) on depth boundary completeness ($\varepsilon_{\mathrm{DBE}}^{\text {comp }}$) and planarity ($\varepsilon_{\mathrm{PE}}^{\text {plan }}$), respectively.

Figure~\ref{fig:ibims_vis} visualizes the OOD performance of the three pre-training strategies. From the results, we observe a significant improvement in the sharpness of depth maps when leveraging geometric and supervised pre-training. Moreover, the localization of depth boundaries is also greatly enhanced.
\subsection{Layer-wise Analysis of MeSa}
\label{sec:magis_cka}
\begin{figure}[t]
  \centering
  \renewcommand{\arraystretch}{0.0} % decrease vertical spacing
  \begin{tabular}{c}
    \begin{subfigure}{0.92\textwidth}
      \centering
      \includegraphics[width=\textwidth]{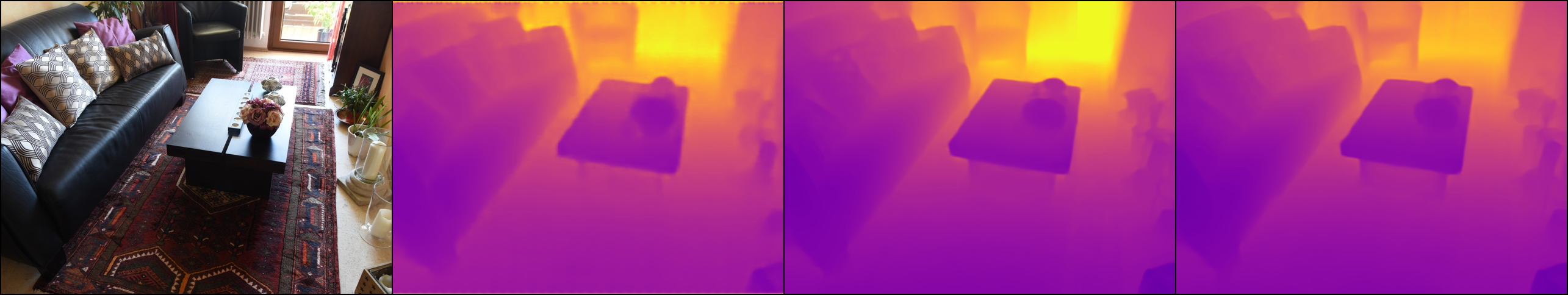}
    \end{subfigure} \\
    \begin{subfigure}{0.92\textwidth}
      \centering
      \includegraphics[width=\textwidth]{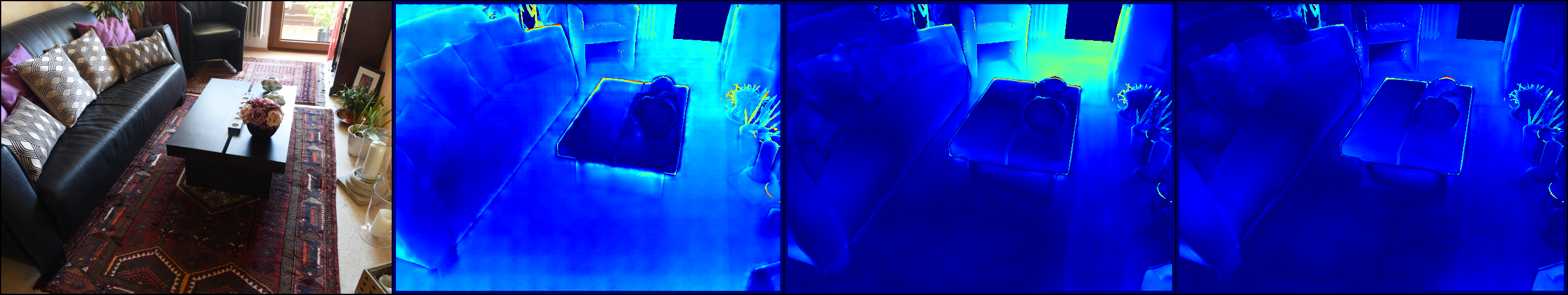}
    \end{subfigure} \\
    \begin{subfigure}{0.92\textwidth}
      \centering
      \includegraphics[width=\textwidth]{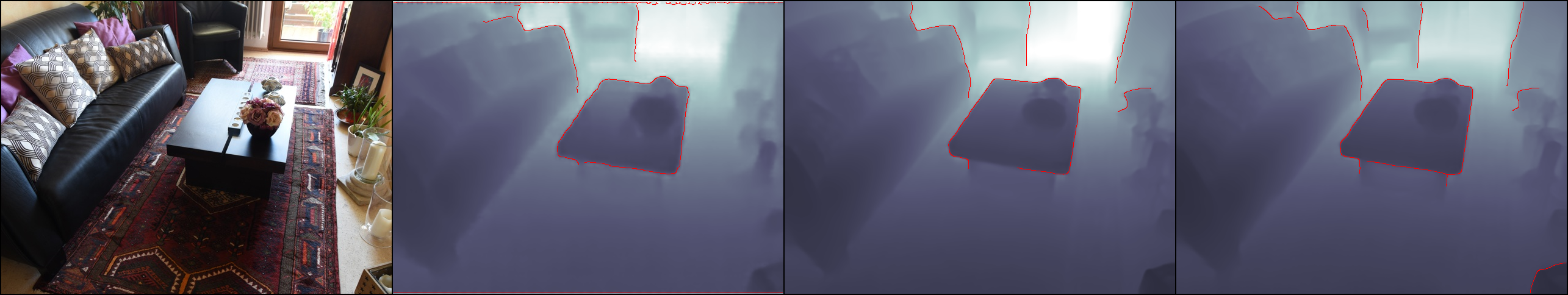}
    \end{subfigure} \\
  \end{tabular}
  \caption{Qualitative comparisons of the OOD performance of the three pre-training strategies on IBims-1. Left$\rightarrow$right: Image, MP, MP+GP, MP+GP+SP (MeSa). \textbf{Top}: predicted depth maps, \textbf{middle}: error maps (blue$\rightarrow$lower error; red$\rightarrow$higher error), \textbf{bottom}: edge maps.}
  \label{fig:ibims_vis}
\end{figure}
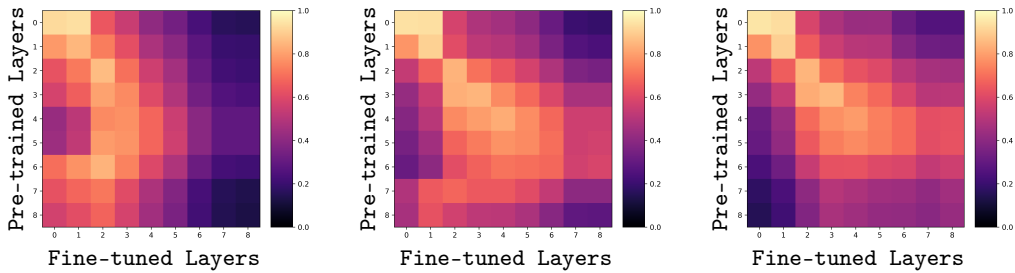
\begin{figure}[t]
    \begin{minipage}{.33\linewidth}
    \centering
        \includegraphics[width=0.95\textwidth]{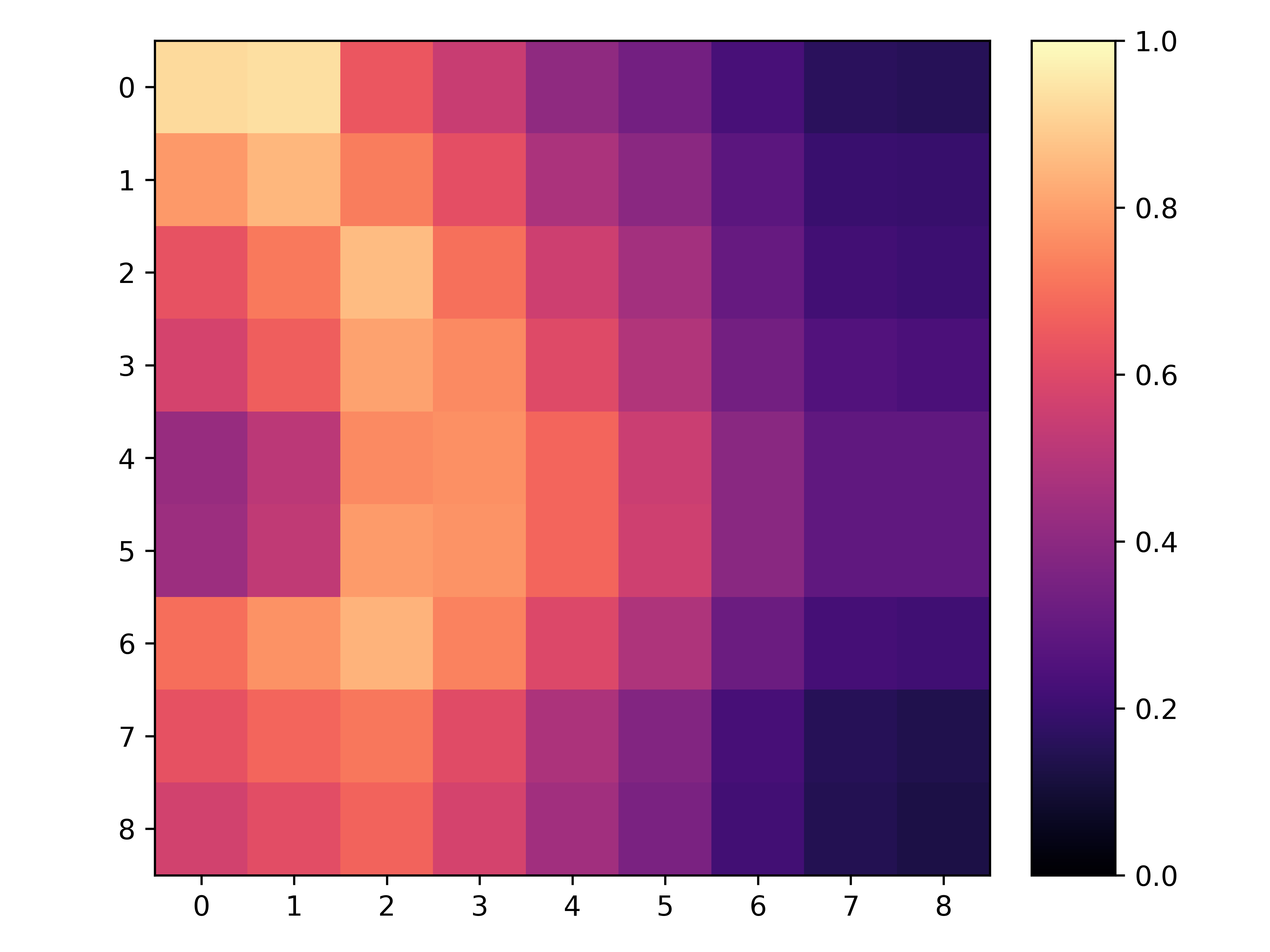}
        \vspace{5pt}
        \begin{picture}(0,0)
        \put(-125,6){\rotatebox{90}{{\footnotesize \texttt{Pre-trained Layers}}}}
        \put(-110,-7){{\footnotesize \texttt{Fine-tuned Layers}}}
        \end{picture}
    \end{minipage}\hfill
    \begin{minipage}{.33\linewidth}
    \centering
        \includegraphics[width=0.95\textwidth]{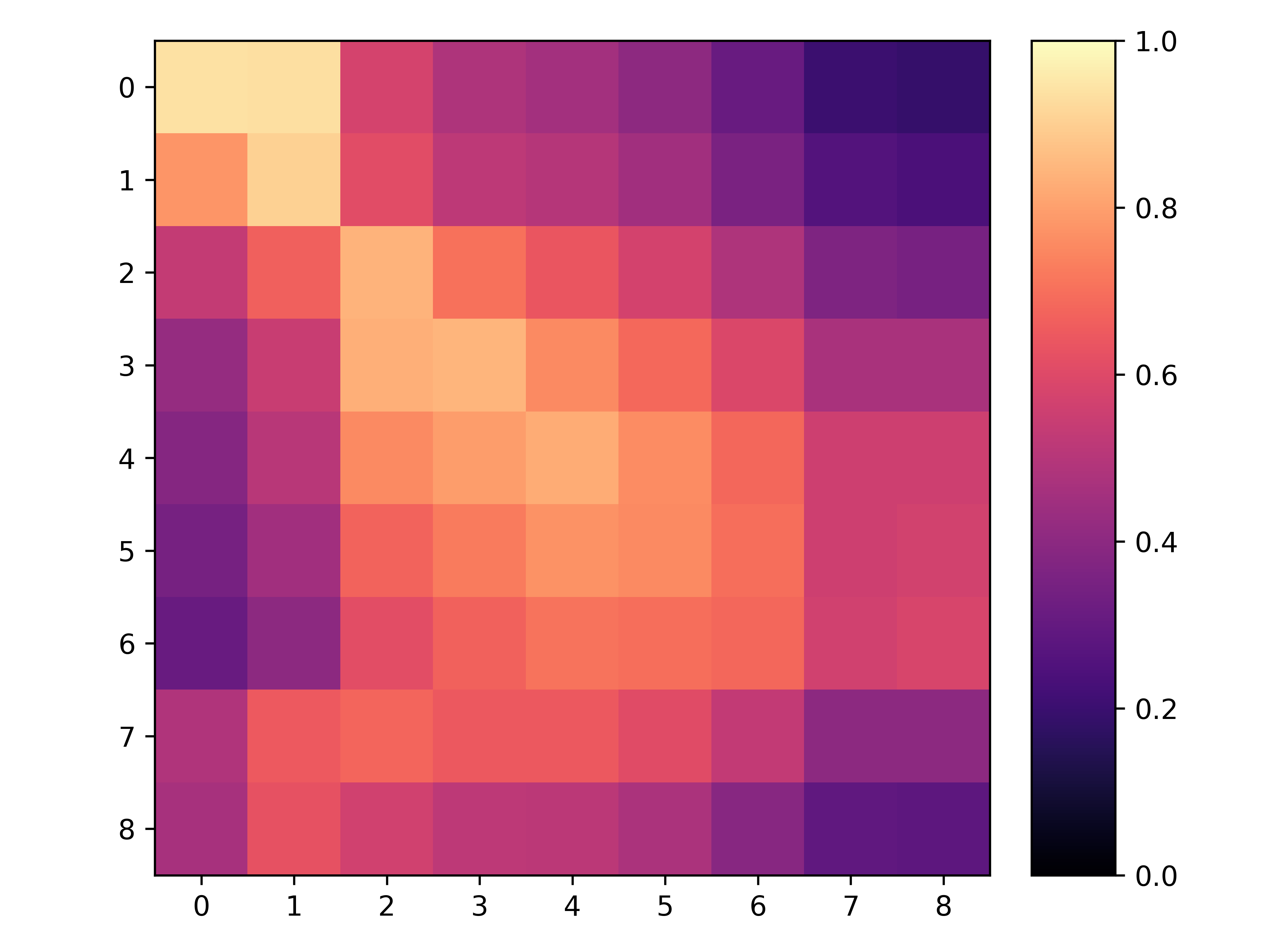}
        \vspace{5pt}
        \begin{picture}(0,0)
        \put(-125,6){\rotatebox{90}{{\footnotesize \texttt{Pre-trained Layers}}}}
        \put(-110,-7){{\footnotesize \texttt{Fine-tuned Layers}}}
        \end{picture}
    \end{minipage}\hfill
    \begin{minipage}{.33\linewidth}
    \centering
        \includegraphics[width=0.95\textwidth]{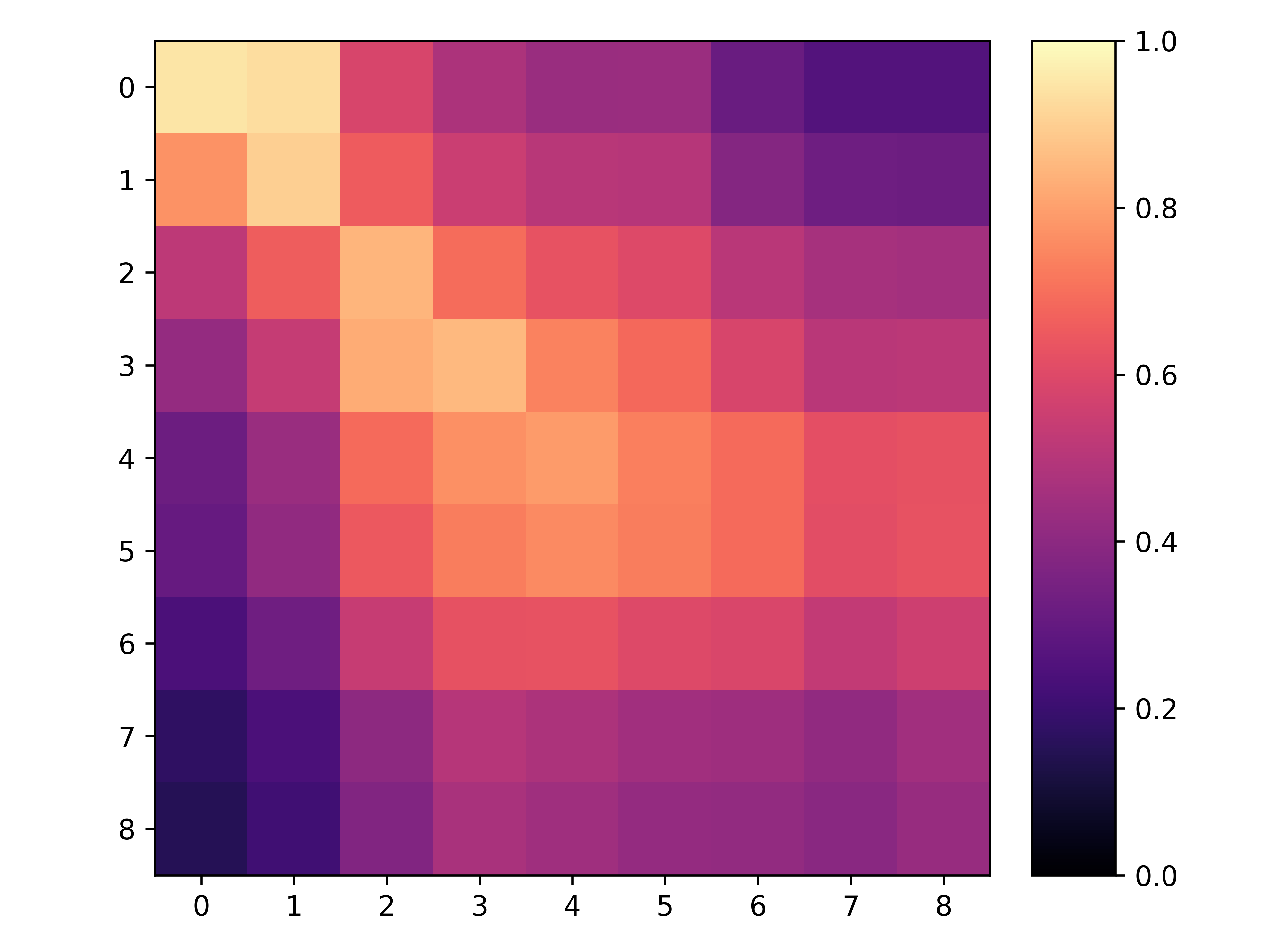}
        \vspace{5pt}
        \begin{picture}(0,0)
        \put(-125,6){\rotatebox{90}{{\footnotesize \texttt{Pre-trained Layers}}}}
        \put(-110,-7){{\footnotesize \texttt{Fine-tuned Layers}}}
        \end{picture}
    \end{minipage}
    \caption{Layer-wise analysis of the three pre-training strategies, comparing the representations of the pre-trained model to the fine-tuned model. MeSa effectively pre-trains the entire network, including the last few layers. Left$\rightarrow$right: MP, MP+GP, MP+GP+SP (MeSa).}
    \label{fig:magis_cka}
\end{figure}
In this section, we analyze layer-wise representations of the MeSa pre-trained network to ascertain that our pre-training strategy indeed produces improved representations for the later layers, thereby successfully overcoming the drawbacks of the SOTA SSL method.

Similar to Section~\ref{sec:sota_cka}, for each layer, we compare the similarity of its pre-trained representation to its fine-tuned representation (Figure~\ref{fig:magis_cka}). As mentioned earlier, when using masked pre-training (MP) alone, the later layers (5-8) are not particularly beneficial for depth estimation since they undergo significant changes during fine-tuning, illustrated by the low similarity values. On the other hand, incorporating geometric pre-training (GP) alongside MP (i.e., MP+GP) allows for more effective representations to be learned deeper into the network, as shown by the higher similarities (lighter colors) until layer 6. However, layers 7-8 are still not optimally utilized. By leveraging all three pre-training strategies (i.e., MP+GP+SP), even these last two layers have comparatively higher similarities. Hence, MeSa effectively pre-trains the entire network, including the last few layers, for downstream depth estimation. The Appendix shows more CKA layer-wise analyses illustrating the efficacy of MeSa pre-training.
\subsection{Impact of Different Pre-training Datasets on Layer-wise Features}
\label{sec:lsun_pretraining}
In this section, we study the impact of different pre-training datasets on the layer-wise pre-trained features. To investigate this, we compare the representations of two networks pre-trained on LSUN and ImageNet respectively. LSUN is less diverse (featuring mostly indoor scenes) in comparison to ImageNet, allowing us to understand the impact of pre-training on these two datasets of varying diversities. To ensure a fair comparison, we use the same size of pre-training dataset (1200K) in both cases. For ImageNet, we use the pre-trained model provided by~\citet{xie2022simmim,xie2022revealing} whereas for LSUN, we pre-train a Swin-v2-L network using SimMIM~\citep{xie2022simmim} on a subset of the LSUN dataset. The subset consists of 300K images each from the bedroom, dining room, living room, and kitchen categories.

Similar to the analysis in Section~\ref{sec:sota_cka}, Figure~\ref{fig:imagenet_vs_lsun_p_p_cka} (in the Appendix) compares the representations of various layers within the pre-trained networks, whereas Figure~\ref{fig:imagenet_vs_lsun_p_ft_cka} compares the representations of the pre-trained networks and the (depth estimation) fine-tuned networks. The analysis in Figure~\ref{fig:imagenet_vs_lsun_p_p_cka} (top row) reveals that the intermediate layers of the LSUN pre-trained network exhibit lower similarities (darker colors) with layer 0 compared to the similarities of the corresponding layers of the ImageNet pre-trained network. Since earlier layers (e.g., layer 0) capture low-level features, this suggests that pre-training with LSUN learns more high-level representations in the middle layers that are beneficial for tasks requiring higher-level reasoning, as illustrated by the superior depth estimation results on NYUv2. This is further corroborated by Figure~\ref{fig:imagenet_vs_lsun_p_ft_cka}, which demonstrates that the LSUN pre-trained features are more effective for the downstream task than the ImageNet pre-trained features, as evidenced by the higher similarities (lighter colors) between the pre-trained and fine-tuned features along the diagonal line for LSUN compared to ImageNet. Since the images in LSUN have a smaller diversity compared to those in ImageNet, we hypothesize that pre-training on LSUN forces the network to reconstruct masked portions of similar images, thereby encouraging it to utilize more layers for higher level reasoning than for RGB reconstruction (Figure~\ref{fig:imagenet_vs_lsun_p_p_cka}).
\subsection{Comparison with SOTA Methods}
\label{sec:sota}
\begin{table}
\caption{Comparison with SOTA depth estimation methods on the NYUv2 dataset. MeSa outperforms previous methods across all metrics.}
\label{table:sota}
\centering
\begin{tabular}{lcccccc}\toprule
Method & RMSE $\downarrow$ & $\delta_1 \uparrow$ & $\delta_2 \uparrow$ & $\delta_3 \uparrow$ & $\operatorname{REL} \downarrow$ & $\log 10 \downarrow$ \\ \midrule
BTS~\citep{kirillov2019panoptic} & 0.392 & 0.885 & 0.978 & 0.995 & 0.110 & 0.047 \\
AdaBins~\citep{bhat2021adabins} & 0.364 & 0.903 & 0.984 & 0.997 & 0.103 & 0.044 \\
DPT~\citep{ranftl2021vision} & 0.357 & 0.904 & 0.988 & 0.998 & 0.110 & 0.045 \\
P3Depth~\citep{patil2022p3depth} & 0.356 & 0.898 & 0.981 & 0.996 & 0.104 & 0.043 \\
NeWCRFs~\citep{yuan2022new} & 0.334 & 0.922 & 0.992 & 0.998 & 0.095 & 0.041 \\
Xie et al.~\citep{xie2022revealing} & 0.287 & 0.949 & 0.994 & \textbf{0.999} & 0.083 & 0.035 \\
AiT~\citep{ning2023all} & 0.275 & 0.954 & 0.994 & \textbf{0.999} & 0.076 & 0.033 \\
ZoeDepth~\citep{bhat2023zoedepth} & 0.270 & 0.955 & \textbf{0.995} & \textbf{0.999} & 0.075 & 0.032 \\
VPD~\citep{zhao2023unleashing} & 0.254 & \textbf{0.964} & \textbf{0.995} & \textbf{0.999} & 0.069 & 0.030 \\ \midrule
\METHOD & \textbf{0.238} & \textbf{0.964} & \textbf{0.995} & \textbf{0.999} & \textbf{0.066} & \textbf{0.029} \\
\bottomrule
\end{tabular}
\end{table}
Table~\ref{table:sota} compares MeSa against state-of-the-art methods on the NYUv2 dataset. As in recent works, we primarily focus on RMSE since performance on the other metrics is highly saturated. Our method is most directly comparable to~\citep{xie2022revealing}, which we surpass by a large margin of 17.1\%. Moreover, even without utilizing any recently proposed techniques, such as soft token and mask augmentation~\citep{ning2023all}, metric bins module~\citep{bhat2023zoedepth} or pre-trained text-to-image diffusion model~\citep{zhao2023unleashing}, our approach also outperforms the most recent methods. Since our method is agnostic to these techniques, it could also be used in conjunction with them to yield further improvements. Overall, we surpass the current best approach~\citep{zhao2023unleashing} by 6.3\%, establishing a new SOTA on the challenging NYUv2 dataset.
\section{Conclusion}
\label{sec:conclusion}
We propose MeSa, a novel pre-training pipeline, to overcome the limitations of the SOTA SSL method, which fails to optimally pre-train the later layers. Via integrating the complementary strengths of masked, geometric, and supervised pre-training, MeSa learns effective layer-wise representations across the entire network. Moreover, MeSa leads to a substantial improvement of 17.1\% on the RMSE compared to the SOTA SSL method and establishes a new state-of-the-art for monocular depth estimation on the challenging NYUv2 dataset.

\bibliography{bibliography}

\begin{thebibliography}{49}
\providecommand{\natexlab}[1]{#1}
\providecommand{\url}[1]{\texttt{#1}}
\expandafter\ifx\csname urlstyle\endcsname\relax
  \providecommand{\doi}[1]{doi: #1}\else
  \providecommand{\doi}{doi: \begingroup \urlstyle{rm}\Url}\fi

\bibitem[Bhat et~al.(2021)Bhat, Alhashim, and Wonka]{bhat2021adabins}
S.~F. Bhat, I.~Alhashim, and P.~Wonka.
\newblock Adabins: Depth estimation using adaptive bins.
\newblock In \emph{Proceedings of the IEEE/CVF Conference on Computer Vision and Pattern Recognition}, pages 4009--4018, 2021.

\bibitem[Bhat et~al.(2023)Bhat, Birkl, Wofk, Wonka, and M{\"u}ller]{bhat2023zoedepth}
S.~F. Bhat, R.~Birkl, D.~Wofk, P.~Wonka, and M.~M{\"u}ller.
\newblock Zoedepth: Zero-shot transfer by combining relative and metric depth.
\newblock \emph{arXiv preprint arXiv:2302.12288}, 2023.

\bibitem[Bian et~al.(2021{\natexlab{a}})Bian, Zhan, Wang, Chin, Shen, and Reid]{bian2021tpami}
J.-W. Bian, H.~Zhan, N.~Wang, T.-J. Chin, C.~Shen, and I.~Reid.
\newblock Auto-rectify network for unsupervised indoor depth estimation.
\newblock \emph{IEEE PAMI}, 2021{\natexlab{a}}.

\bibitem[Bian et~al.(2021{\natexlab{b}})Bian, Zhan, Wang, Li, Zhang, Shen, Cheng, and Reid]{bian2021ijcv}
J.-W. Bian, H.~Zhan, N.~Wang, Z.~Li, L.~Zhang, C.~Shen, M.-M. Cheng, and I.~Reid.
\newblock Unsupervised scale-consistent depth learning from video.
\newblock \emph{IJCV}, 2021{\natexlab{b}}.

\bibitem[Cao et~al.(2020)Cao, Xie, Liu, Lin, Zhang, and Hu]{cao2020pic}
Y.~Cao, Z.~Xie, B.~Liu, Y.~Lin, Z.~Zhang, and H.~Hu.
\newblock Parametric instance classification for unsupervised visual feature learning.
\newblock \emph{Advances in Neural Information Processing Systems}, 33, 2020.

\bibitem[Chen et~al.(2020)Chen, Kornblith, Norouzi, and Hinton]{chen2020simclr}
T.~Chen, S.~Kornblith, M.~Norouzi, and G.~Hinton.
\newblock A simple framework for contrastive learning of visual representations.
\newblock \emph{ICML}, 2020.

\bibitem[Chen et~al.(2019)Chen, Schmid, and Sminchisescu]{chen2019self}
Y.~Chen, C.~Schmid, and C.~Sminchisescu.
\newblock Self-supervised learning with geometric constraints in monocular video: Connecting flow, depth, and camera.
\newblock In \emph{ICCV}, pages 7063--7072, 2019.

\bibitem[Devlin et~al.(2018)Devlin, Chang, Lee, and Toutanova]{devlin2018bert}
J.~Devlin, M.-W. Chang, K.~Lee, and K.~Toutanova.
\newblock Bert: Pre-training of deep bidirectional transformers for language understanding.
\newblock \emph{arXiv preprint arXiv:1810.04805}, 2018.

\bibitem[Doersch et~al.(2015)Doersch, Gupta, and Efros]{doersch2015context}
C.~Doersch, A.~Gupta, and A.~A. Efros.
\newblock Unsupervised visual representation learning by context prediction.
\newblock In \emph{ICCV}, pages 1422--1430, 2015.

\bibitem[Eigen et~al.(2014)Eigen, Puhrsch, and Fergus]{eigen2014depth}
D.~Eigen, C.~Puhrsch, and R.~Fergus.
\newblock Depth map prediction from a single image using a multi-scale deep network.
\newblock In \emph{NeurIPS}, 2014.

\bibitem[Fu et~al.(2018)Fu, Gong, Wang, Batmanghelich, and Tao]{fu2018deep}
H.~Fu, M.~Gong, C.~Wang, K.~Batmanghelich, and D.~Tao.
\newblock Deep ordinal regression network for monocular depth estimation.
\newblock In \emph{CVPR}, pages 2002--2011, 2018.

\bibitem[Garg et~al.(2016)Garg, BG, Carneiro, and Reid]{garg2016unsupervised}
R.~Garg, V.~K. BG, G.~Carneiro, and I.~Reid.
\newblock Unsupervised cnn for single view depth estimation: Geometry to the rescue.
\newblock In \emph{ECCV}. Springer, 2016.

\bibitem[Gidaris et~al.(2018)Gidaris, Singh, and Komodakis]{gidaris2018rotation}
S.~Gidaris, P.~Singh, and N.~Komodakis.
\newblock Unsupervised representation learning by predicting image rotations.
\newblock \emph{arXiv preprint arXiv:1803.07728}, 2018.

\bibitem[Godard et~al.(2017)Godard, Mac~Aodha, and Brostow]{godard2017unsupervised}
C.~Godard, O.~Mac~Aodha, and G.~J. Brostow.
\newblock Unsupervised monocular depth estimation with left-right consistency.
\newblock In \emph{CVPR}, 2017.

\bibitem[Godard et~al.(2019)Godard, {Mac Aodha}, Firman, and Brostow]{monodepth2}
C.~Godard, O.~{Mac Aodha}, M.~Firman, and G.~J. Brostow.
\newblock Digging into self-supervised monocular depth prediction.
\newblock In \emph{ICCV}, 2019.

\bibitem[Grill et~al.(2020)Grill, Strub, Altch{\'e}, Tallec, Richemond, Buchatskaya, Doersch, Avila~Pires, Guo, Gheshlaghi~Azar, et~al.]{grill2020byol}
J.-B. Grill, F.~Strub, F.~Altch{\'e}, C.~Tallec, P.~Richemond, E.~Buchatskaya, C.~Doersch, B.~Avila~Pires, Z.~Guo, M.~Gheshlaghi~Azar, et~al.
\newblock Bootstrap your own latent-a new approach to self-supervised learning.
\newblock \emph{Advances in Neural Information Processing Systems}, 33, 2020.

\bibitem[Guizilini et~al.(2020)Guizilini, Ambrus, Pillai, Raventos, and Gaidon]{packnet}
V.~Guizilini, R.~Ambrus, S.~Pillai, A.~Raventos, and A.~Gaidon.
\newblock 3d packing for self-supervised monocular depth estimation.
\newblock In \emph{CVPR}, 2020.

\bibitem[Guo et~al.(2022)Guo, Islam, Gotway, and Liang]{guo2022discriminative}
Z.~Guo, N.~U. Islam, M.~B. Gotway, and J.~Liang.
\newblock Discriminative, restorative, and adversarial learning: Stepwise incremental pretraining.
\newblock In \emph{Domain Adaptation and Representation Transfer: 4th MICCAI Workshop, DART 2022, Held in Conjunction with MICCAI 2022, Singapore, September 22, 2022, Proceedings}, pages 66--76. Springer, 2022.

\bibitem[He et~al.(2016)He, Zhang, Ren, and Sun]{he2016deep}
K.~He, X.~Zhang, S.~Ren, and J.~Sun.
\newblock Deep residual learning for image recognition.
\newblock In \emph{CVPR}, pages 770--778, 2016.

\bibitem[He et~al.(2020)He, Fan, Wu, Xie, and Girshick]{he2019moco}
K.~He, H.~Fan, Y.~Wu, S.~Xie, and R.~Girshick.
\newblock Momentum contrast for unsupervised visual representation learning.
\newblock \emph{CVPR}, 2020.

\bibitem[He et~al.(2022)He, Chen, Xie, Li, Doll{\'a}r, and Girshick]{he2022masked}
K.~He, X.~Chen, S.~Xie, Y.~Li, P.~Doll{\'a}r, and R.~Girshick.
\newblock Masked autoencoders are scalable vision learners.
\newblock In \emph{Proceedings of the IEEE/CVF Conference on Computer Vision and Pattern Recognition}, pages 16000--16009, 2022.

\bibitem[Jaderberg et~al.(2015)Jaderberg, Simonyan, Zisserman, et~al.]{jaderberg2015stn}
M.~Jaderberg, K.~Simonyan, A.~Zisserman, et~al.
\newblock Spatial transformer networks.
\newblock In \emph{NeurIPS}, 2015.

\bibitem[Kirillov et~al.(2019)Kirillov, Girshick, He, and Doll{\'a}r]{kirillov2019panoptic}
A.~Kirillov, R.~Girshick, K.~He, and P.~Doll{\'a}r.
\newblock Panoptic feature pyramid networks.
\newblock In \emph{Proceedings of the IEEE/CVF conference on computer vision and pattern recognition}, pages 6399--6408, 2019.

\bibitem[Kornblith et~al.(2019)Kornblith, Norouzi, Lee, and Hinton]{kornblith2019similarity}
S.~Kornblith, M.~Norouzi, H.~Lee, and G.~Hinton.
\newblock Similarity of neural network representations revisited.
\newblock In \emph{International Conference on Machine Learning}, pages 3519--3529. PMLR, 2019.

\bibitem[Laina et~al.(2016)Laina, Rupprecht, Belagiannis, Tombari, and Navab]{laina2016deeper}
I.~Laina, C.~Rupprecht, V.~Belagiannis, F.~Tombari, and N.~Navab.
\newblock Deeper depth prediction with fully convolutional residual networks.
\newblock In \emph{3DV}, 2016.

\bibitem[Liu et~al.(2019)Liu, Ott, Goyal, Du, Joshi, Chen, Levy, Lewis, Zettlemoyer, and Stoyanov]{liu2019roberta}
Y.~Liu, M.~Ott, N.~Goyal, J.~Du, M.~Joshi, D.~Chen, O.~Levy, M.~Lewis, L.~Zettlemoyer, and V.~Stoyanov.
\newblock Roberta: A robustly optimized bert pretraining approach.
\newblock \emph{arXiv preprint arXiv:1907.11692}, 2019.

\bibitem[Mahjourian et~al.(2018)Mahjourian, Wicke, and Angelova]{mahjourian2018unsupervised}
R.~Mahjourian, M.~Wicke, and A.~Angelova.
\newblock Unsupervised learning of depth and ego-motion from monocular video using 3d geometric constraints.
\newblock In \emph{CVPR}, 2018.

\bibitem[Mayer et~al.(2018)Mayer, Ilg, Fischer, Hazirbas, Cremers, Dosovitskiy, and Brox]{mayer2018makes}
N.~Mayer, E.~Ilg, P.~Fischer, C.~Hazirbas, D.~Cremers, A.~Dosovitskiy, and T.~Brox.
\newblock What makes good synthetic training data for learning disparity and optical flow estimation?
\newblock \emph{International Journal of Computer Vision}, 126:\penalty0 942--960, 2018.

\bibitem[Neyshabur et~al.(2020)Neyshabur, Sedghi, and Zhang]{neyshabur2020being}
B.~Neyshabur, H.~Sedghi, and C.~Zhang.
\newblock What is being transferred in transfer learning?
\newblock \emph{Advances in neural information processing systems}, 33:\penalty0 512--523, 2020.

\bibitem[Ning et~al.(2023)Ning, Li, Zhang, Geng, Dai, He, and Hu]{ning2023all}
J.~Ning, C.~Li, Z.~Zhang, Z.~Geng, Q.~Dai, K.~He, and H.~Hu.
\newblock All in tokens: Unifying output space of visual tasks via soft token.
\newblock \emph{arXiv preprint arXiv:2301.02229}, 2023.

\bibitem[Pasad et~al.(2021)Pasad, Chou, and Livescu]{pasad2021layer}
A.~Pasad, J.-C. Chou, and K.~Livescu.
\newblock Layer-wise analysis of a self-supervised speech representation model.
\newblock In \emph{2021 IEEE Automatic Speech Recognition and Understanding Workshop (ASRU)}, pages 914--921. IEEE, 2021.

\bibitem[Pasad et~al.(2023)Pasad, Shi, and Livescu]{pasad2023comparative}
A.~Pasad, B.~Shi, and K.~Livescu.
\newblock Comparative layer-wise analysis of self-supervised speech models.
\newblock In \emph{ICASSP 2023-2023 IEEE International Conference on Acoustics, Speech and Signal Processing (ICASSP)}, pages 1--5. IEEE, 2023.

\bibitem[Patil et~al.(2022)Patil, Sakaridis, Liniger, and Van~Gool]{patil2022p3depth}
V.~Patil, C.~Sakaridis, A.~Liniger, and L.~Van~Gool.
\newblock P3depth: Monocular depth estimation with a piecewise planarity prior.
\newblock In \emph{Proceedings of the IEEE/CVF Conference on Computer Vision and Pattern Recognition}, pages 1610--1621, 2022.

\bibitem[Ranftl et~al.(2020)Ranftl, Lasinger, Hafner, Schindler, and Koltun]{ranftl2020towards}
R.~Ranftl, K.~Lasinger, D.~Hafner, K.~Schindler, and V.~Koltun.
\newblock Towards robust monocular depth estimation: Mixing datasets for zero-shot cross-dataset transfer.
\newblock \emph{IEEE PAMI}, 2020.

\bibitem[Ranftl et~al.(2021)Ranftl, Bochkovskiy, and Koltun]{ranftl2021vision}
R.~Ranftl, A.~Bochkovskiy, and V.~Koltun.
\newblock Vision transformers for dense prediction.
\newblock In \emph{Proceedings of the IEEE/CVF International Conference on Computer Vision}, pages 12179--12188, 2021.

\bibitem[Silberman et~al.(2012)Silberman, Hoiem, Kohli, and Fergus]{silberman2012indoor}
N.~Silberman, D.~Hoiem, P.~Kohli, and R.~Fergus.
\newblock Indoor segmentation and support inference from rgbd images.
\newblock In \emph{ECCV}, 2012.

\bibitem[Sun et~al.(2022)Sun, Bian, Zhan, Yin, Reid, and Shen]{sun2022sc}
L.~Sun, J.-W. Bian, H.~Zhan, W.~Yin, I.~Reid, and C.~Shen.
\newblock Sc-depthv3: Robust self-supervised monocular depth estimation for dynamic scenes.
\newblock \emph{arXiv preprint arXiv:2211.03660}, 2022.

\bibitem[Wang et~al.(2004)Wang, Bovik, Sheikh, Simoncelli, et~al.]{wang2004image}
Z.~Wang, A.~C. Bovik, H.~R. Sheikh, E.~P. Simoncelli, et~al.
\newblock {Image Quality Assessment}: from error visibility to structural similarity.
\newblock \emph{IEEE TIP}, 13\penalty0 (4), 2004.

\bibitem[Wu et~al.(2018)Wu, Xiong, Yu, and Lin]{wu2018memorybank}
Z.~Wu, Y.~Xiong, S.~X. Yu, and D.~Lin.
\newblock Unsupervised feature learning via non-parametric instance discrimination.
\newblock In \emph{CVPR}, pages 3733--3742, 2018.

\bibitem[Xie et~al.(2021)Xie, Lin, Zhang, Cao, Lin, and Hu]{xie2021pixpro}
Z.~Xie, Y.~Lin, Z.~Zhang, Y.~Cao, S.~Lin, and H.~Hu.
\newblock Propagate yourself: Exploring pixel-level consistency for unsupervised visual representation learning.
\newblock In \emph{Proceedings of the IEEE/CVF Conference on Computer Vision and Pattern Recognition}, pages 16684--16693, 2021.

\bibitem[Xie et~al.(2022{\natexlab{a}})Xie, Geng, Hu, Zhang, Hu, and Cao]{xie2022revealing}
Z.~Xie, Z.~Geng, J.~Hu, Z.~Zhang, H.~Hu, and Y.~Cao.
\newblock Revealing the dark secrets of masked image modeling.
\newblock \emph{arXiv preprint arXiv:2205.13543}, 2022{\natexlab{a}}.

\bibitem[Xie et~al.(2022{\natexlab{b}})Xie, Zhang, Cao, Lin, Bao, Yao, Dai, and Hu]{xie2022simmim}
Z.~Xie, Z.~Zhang, Y.~Cao, Y.~Lin, J.~Bao, Z.~Yao, Q.~Dai, and H.~Hu.
\newblock Simmim: A simple framework for masked image modeling.
\newblock In \emph{Proceedings of the IEEE/CVF Conference on Computer Vision and Pattern Recognition}, pages 9653--9663, 2022{\natexlab{b}}.

\bibitem[Yin et~al.(2021)Yin, Zhang, Wang, Niklaus, Mai, Chen, and Shen]{yin2021learning}
W.~Yin, J.~Zhang, O.~Wang, S.~Niklaus, L.~Mai, S.~Chen, and C.~Shen.
\newblock Learning to recover 3d scene shape from a single image.
\newblock In \emph{CVPR}, pages 204--213, 2021.

\bibitem[Yin and Shi(2018)]{yin2018geonet}
Z.~Yin and J.~Shi.
\newblock {GeoNet}: Unsupervised learning of dense depth, optical flow and camera pose.
\newblock In \emph{CVPR}, 2018.

\bibitem[Yu et~al.(2015)Yu, Seff, Zhang, Song, Funkhouser, and Xiao]{yu2015lsun}
F.~Yu, A.~Seff, Y.~Zhang, S.~Song, T.~Funkhouser, and J.~Xiao.
\newblock Lsun: Construction of a large-scale image dataset using deep learning with humans in the loop.
\newblock \emph{arXiv preprint arXiv:1506.03365}, 2015.

\bibitem[Yuan et~al.(2022)Yuan, Gu, Dai, Zhu, and Tan]{yuan2022new}
W.~Yuan, X.~Gu, Z.~Dai, S.~Zhu, and P.~Tan.
\newblock New crfs: Neural window fully-connected crfs for monocular depth estimation.
\newblock \emph{arXiv preprint arXiv:2203.01502}, 2022.

\bibitem[Zhang et~al.(2016)Zhang, Isola, and Efros]{zhang2016colorful}
R.~Zhang, P.~Isola, and A.~A. Efros.
\newblock Colorful image colorization.
\newblock In \emph{Computer Vision--ECCV 2016: 14th European Conference, Amsterdam, The Netherlands, October 11-14, 2016, Proceedings, Part III 14}, pages 649--666. Springer, 2016.

\bibitem[Zhao et~al.(2023)Zhao, Rao, Liu, Liu, Zhou, and Lu]{zhao2023unleashing}
W.~Zhao, Y.~Rao, Z.~Liu, B.~Liu, J.~Zhou, and J.~Lu.
\newblock Unleashing text-to-image diffusion models for visual perception.
\newblock \emph{arXiv preprint arXiv:2303.02153}, 2023.

\bibitem[Zhou et~al.(2017)Zhou, Brown, Snavely, and Lowe]{zhou2017unsupervised}
T.~Zhou, M.~Brown, N.~Snavely, and D.~G. Lowe.
\newblock Unsupervised learning of depth and ego-motion from video.
\newblock In \emph{CVPR}, 2017.

\end{thebibliography}
\bibliographystyle{abbrvnat}

\newpage
\renewcommand{\thesection}{\Alph{section}}
\setcounter{section}{0}
\section{NYUv2 Visualizations}
\begin{figure}[!h]
  \centering
  \renewcommand{\arraystretch}{0.0} % decrease vertical spacing
  \begin{tabular}{c}
    \begin{subfigure}{0.93\textwidth}
      \centering
      \includegraphics[width=\textwidth]{Images/NYUv2/kitchen_rgb_00133_pred_depths.png}
    \end{subfigure} \\
    \vspace{2pt}
    \begin{subfigure}{0.93\textwidth}
      \centering
      \includegraphics[width=\textwidth]{Images/NYUv2/kitchen_rgb_00133_abs_rel_err_depths.png}
    \end{subfigure} \\
    \begin{subfigure}{0.93\textwidth}
      \centering
      \includegraphics[width=\textwidth]{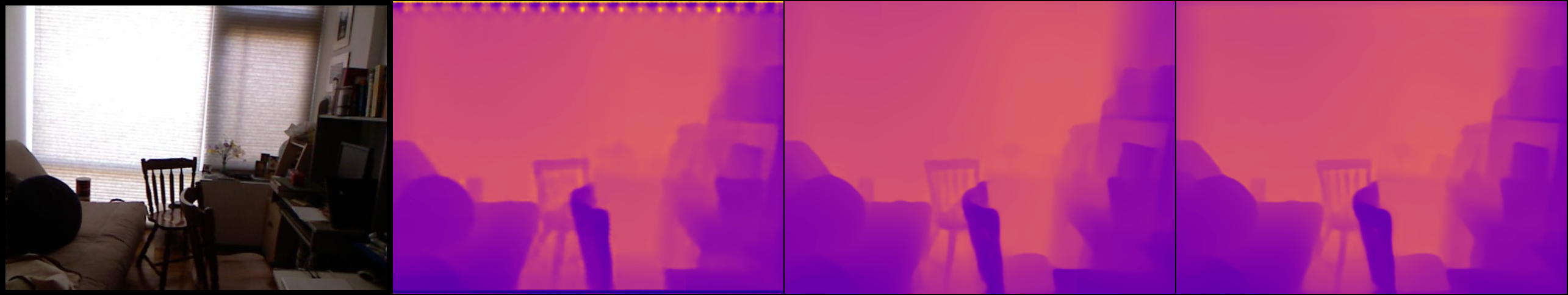}
    \end{subfigure} \\
    \vspace{2pt}
    \begin{subfigure}{0.93\textwidth}
      \centering
      \includegraphics[width=\textwidth]{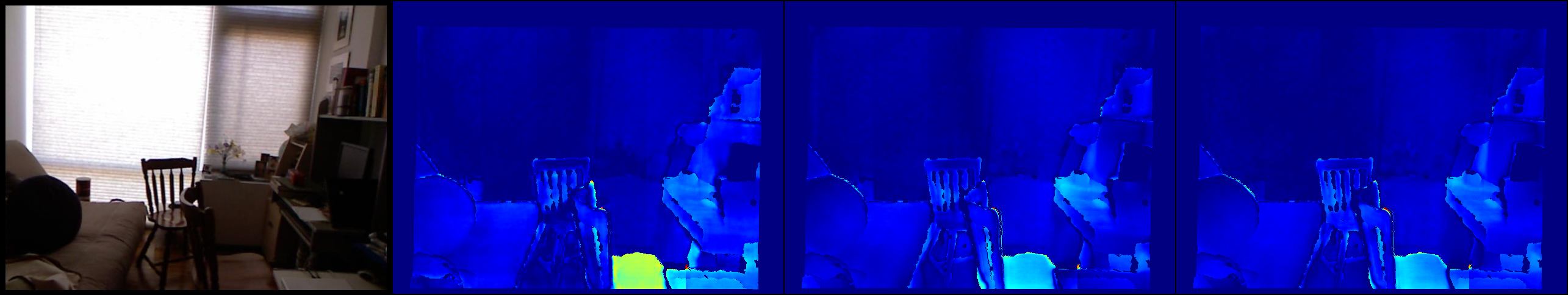}
    \end{subfigure} \\
    \begin{subfigure}{0.93\textwidth}
      \centering
      \includegraphics[width=\textwidth]{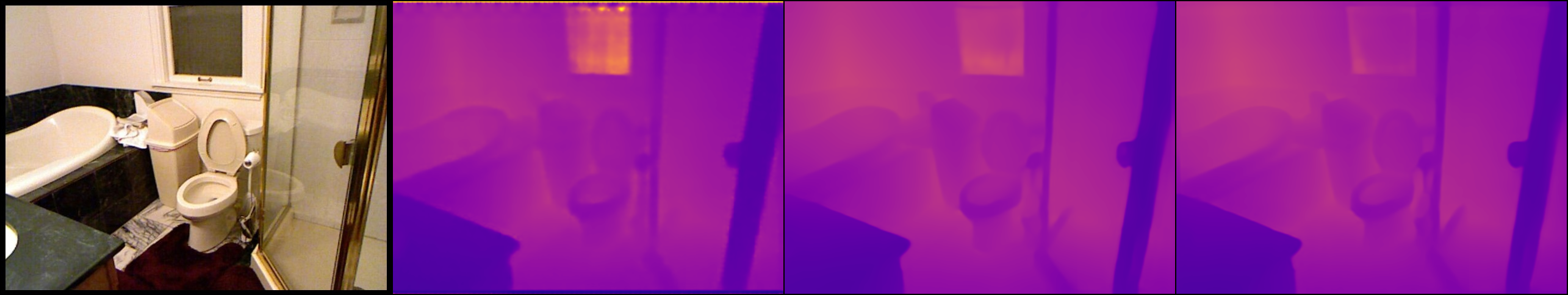}
    \end{subfigure} \\
    \vspace{2pt}
    \begin{subfigure}{0.93\textwidth}
      \centering
      \includegraphics[width=\textwidth]{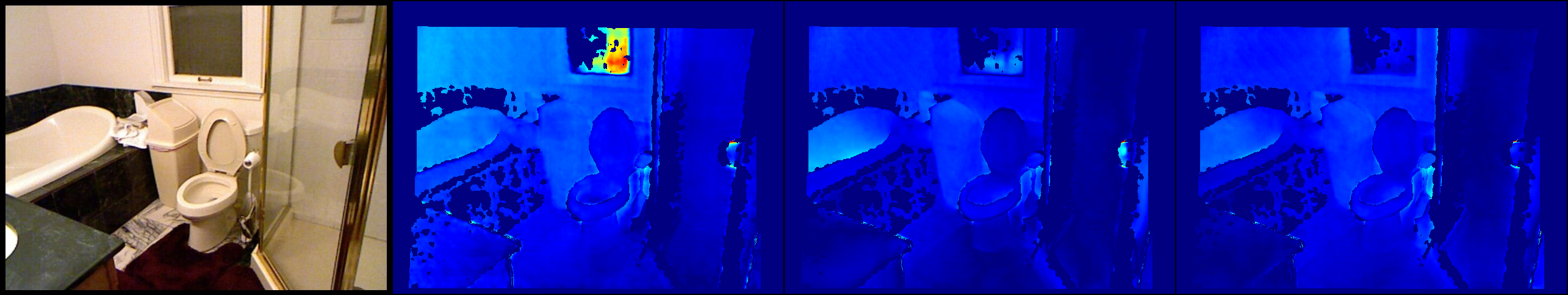}
    \end{subfigure} \\
    \begin{subfigure}{0.93\textwidth}
      \centering
      \includegraphics[width=\textwidth]{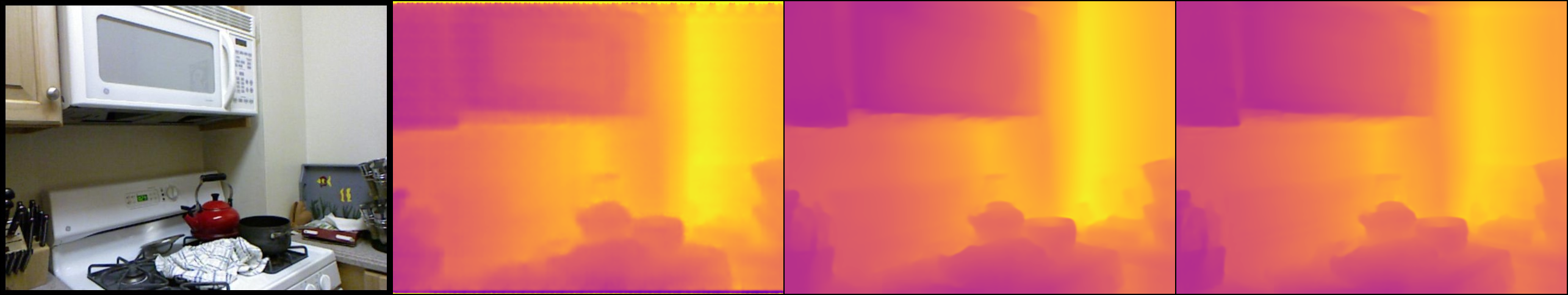}
    \end{subfigure} \\
    \begin{subfigure}{0.93\textwidth}
      \centering
      \includegraphics[width=\textwidth]{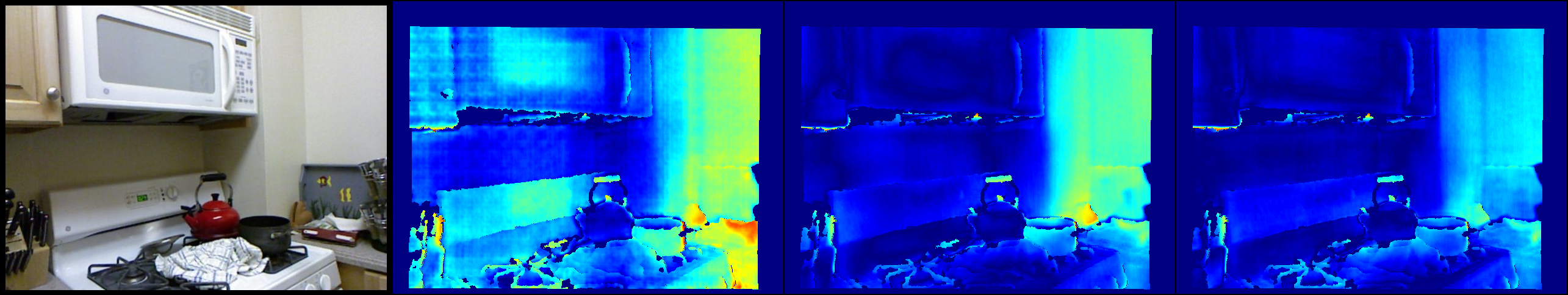}
    \end{subfigure} \\
  \end{tabular}
  \caption{Qualitative comparisons of the three pre-training strategies on NYUv2. Left$\rightarrow$right: Image, MP, MP+GP, MP+GP+SP (MeSa). \textbf{Top}: predicted depth maps, \textbf{bottom}: error maps (blue$\rightarrow$lower error; red$\rightarrow$higher error).}
  \label{fig:nyu_vis_supp}
\end{figure}

Figure~\ref{fig:nyu_vis_supp} shows more visualizations of the three pre-training strategies on the NYUv2 dataset.
\section{IBims-1 Visualizations}
\begin{figure}[!h]
  \centering
  \renewcommand{\arraystretch}{0.0} % decrease vertical spacing
  \begin{tabular}{c}
    \begin{subfigure}{0.77\textwidth}
      \centering
      \includegraphics[width=\textwidth]{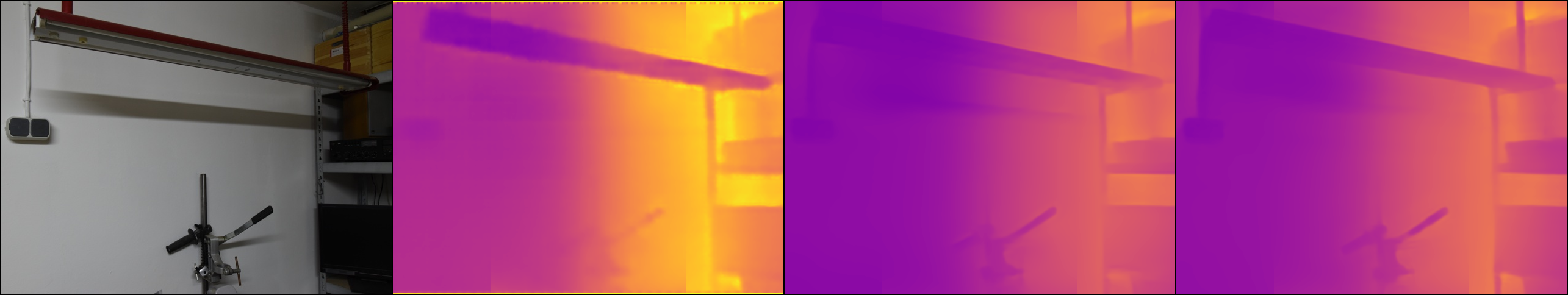}
    \end{subfigure} \\
    \begin{subfigure}{0.77\textwidth}
      \centering
      \includegraphics[width=\textwidth]{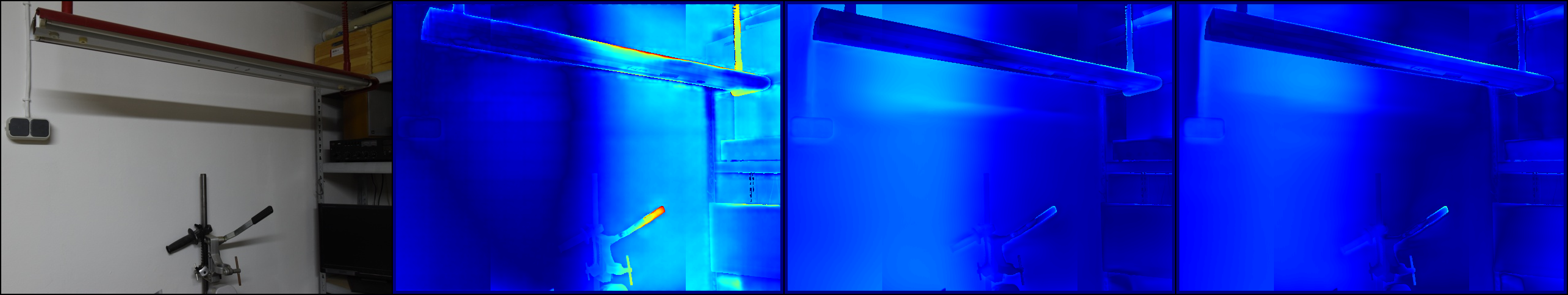}
    \end{subfigure} \\
    \vspace{2pt}
    \begin{subfigure}{0.77\textwidth}
      \centering
      \includegraphics[width=\textwidth]{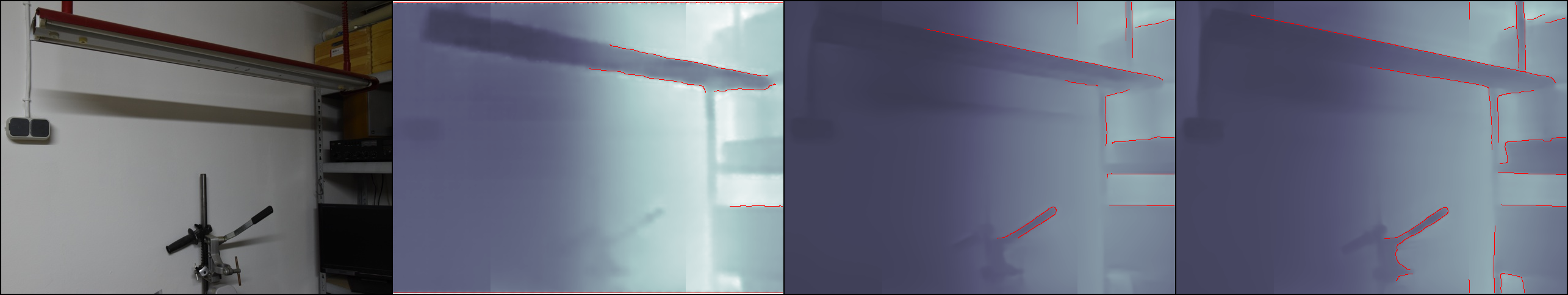}
    \end{subfigure} \\
    \begin{subfigure}{0.77\textwidth}
      \centering
      \includegraphics[width=\textwidth]{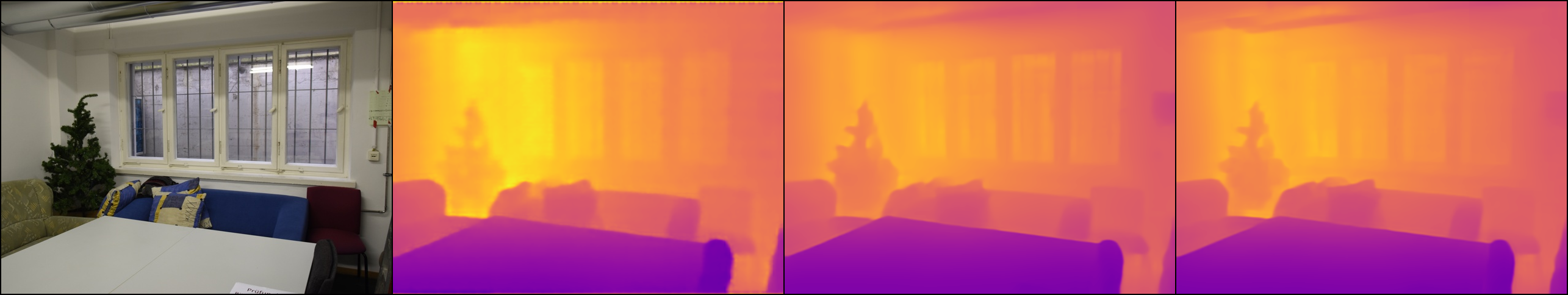}
    \end{subfigure} \\
    \begin{subfigure}{0.77\textwidth}
      \centering
      \includegraphics[width=\textwidth]{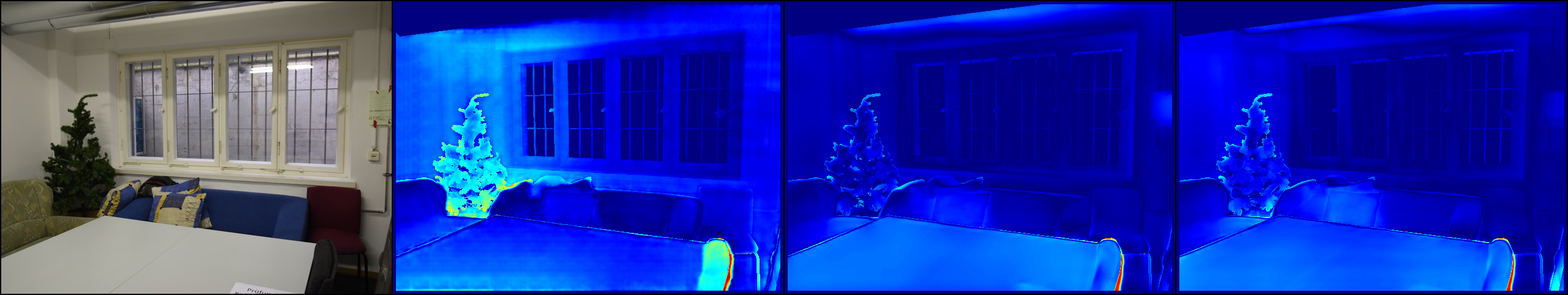}
    \end{subfigure} \\
    \vspace{2pt}
    \begin{subfigure}{0.77\textwidth}
      \centering
      \includegraphics[width=\textwidth]{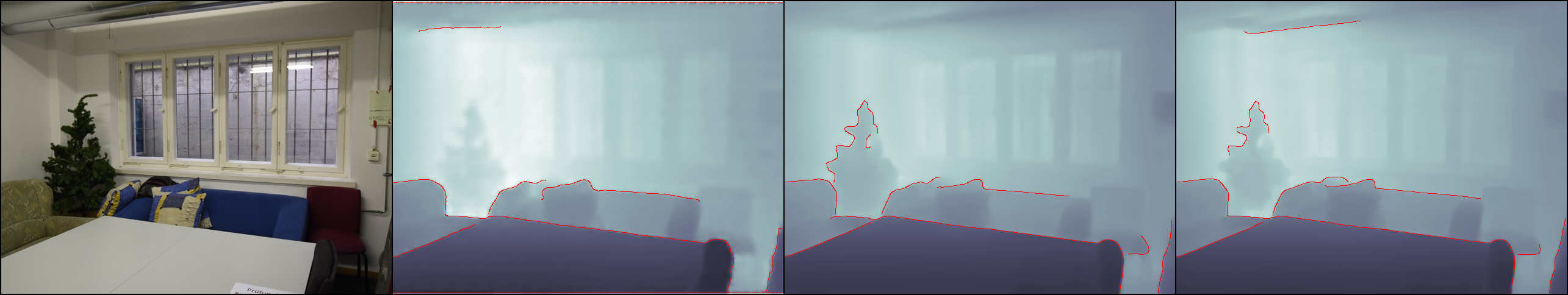}
    \end{subfigure} \\
    \begin{subfigure}{0.77\textwidth}
      \centering
      \includegraphics[width=\textwidth]{Images/IBims/livingroom_14_pred_depths.png}
    \end{subfigure} \\
    \begin{subfigure}{0.77\textwidth}
      \centering
      \includegraphics[width=\textwidth]{Images/IBims/livingroom_14_abs_rel_err_depths.png}
    \end{subfigure} \\
    \begin{subfigure}{0.77\textwidth}
      \centering
      \includegraphics[width=\textwidth]{Images/IBims/livingroom_14_pred_edges.png}
    \end{subfigure} \\
  \end{tabular}
  \caption{Qualitative comparisons of the OOD performance of the three pre-training strategies on IBims-1. Left$\rightarrow$right: Image, MP, MP+GP, MP+GP+SP (MeSa). \textbf{Top}: predicted depth maps, \textbf{middle}: error maps (blue$\rightarrow$lower error; red$\rightarrow$higher error), \textbf{bottom}: edge maps.}
  \label{fig:ibims_vis_supp}
\end{figure}

Figure~\ref{fig:ibims_vis_supp} shows more visualizations of the three pre-training strategies in the OOD setting on the IBims-1 dataset.
\section{CKA Layers}
Similar to previous studies~\citep{xie2022revealing,neyshabur2020being}, we adopt the CKA (centered kernel alignment)~\citep{kornblith2019similarity} metric to compare the representations learnt via the various pre-training methods. The CKA similarity values range from 0 to 1, where higher values indicate greater similarity between the representations. In our investigation, we examine the layer-wise representations of nine layers within the Swin-v2-L architecture. For convenience, we refer to them as layers 0-8, with the exact layers detailed in Table~\ref{table:cka_layers}.
\begin{table}[!h]
\caption{Nine Swin-v2-L layers used for CKA analysis.}
\label{table:cka_layers}
\centering
\begin{tabular}{ll}\toprule
Identifier & Layer in Swin-v2-L \\ \midrule
Layer 0 & layers.0.downsample.norm \\
Layer 1 & layers.1.downsample.norm \\
Layer 2 & layers.2.blocks.3.norm1 \\
Layer 3 & layers.2.blocks.6.norm1 \\
Layer 4 & layers.2.blocks.9.norm1 \\
Layer 5 & layers.2.blocks.12.norm1 \\
Layer 6 & layers.2.blocks.15.norm1 \\
Layer 7 & layers.2.downsample.norm \\
Layer 8 & norm3 \\
\bottomrule
\end{tabular}
\end{table}
\section{MeSa Layer-wise Analysis}
\begin{figure}[!h]
  \centering
    \begin{minipage}{.33\linewidth}
    \centering
        \includegraphics[width=0.95\textwidth]{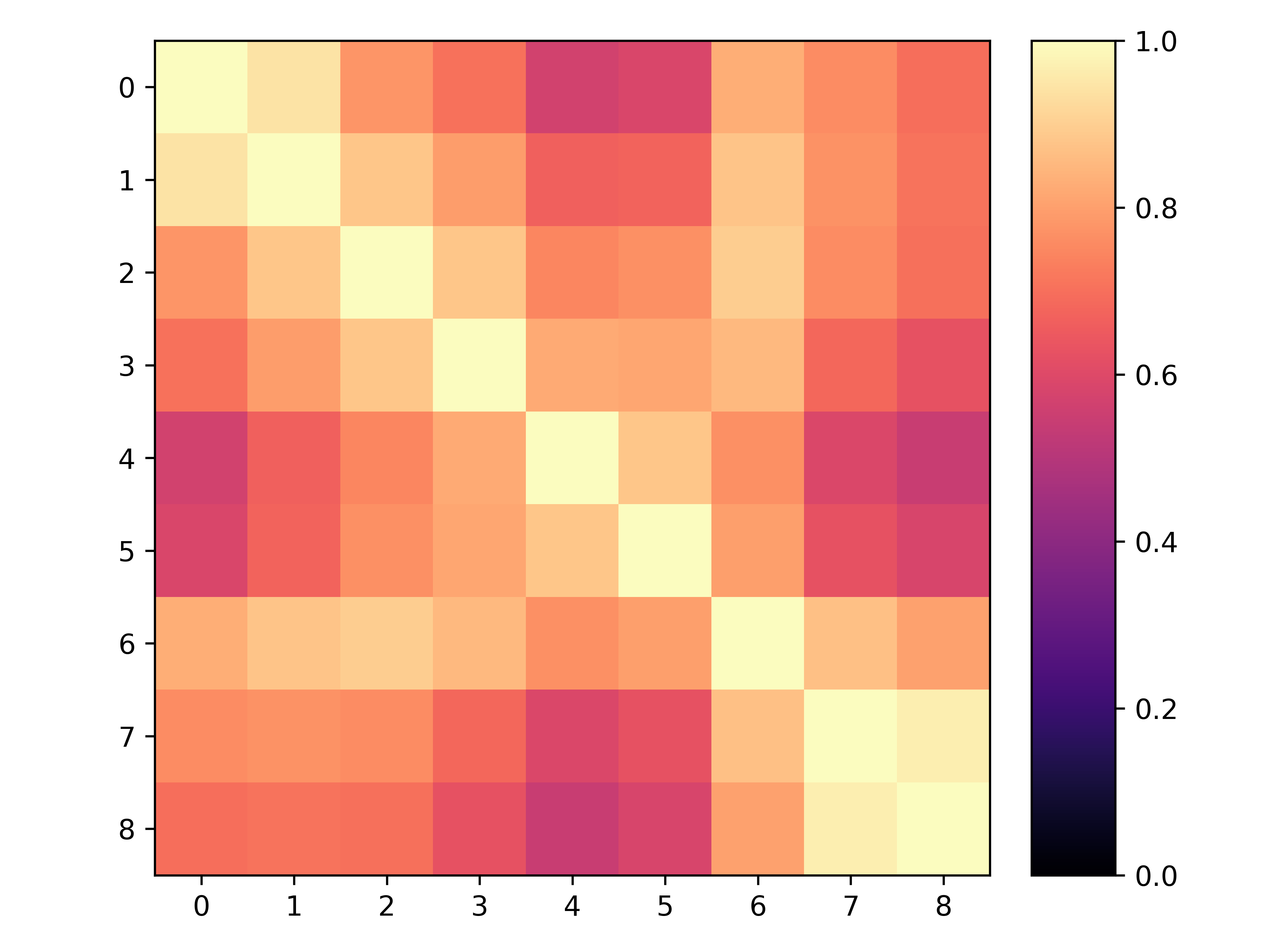}
        \vspace{5pt}
        \begin{picture}(0,0)
        \put(-125,6){\rotatebox{90}{{\footnotesize \texttt{Pre-trained Layers}}}}
        \put(-112,-7){{\footnotesize \texttt{Pre-trained Layers}}}
        \end{picture}
    \end{minipage}\hfill
    \begin{minipage}{.33\linewidth}
    \centering
        \includegraphics[width=0.95\textwidth]{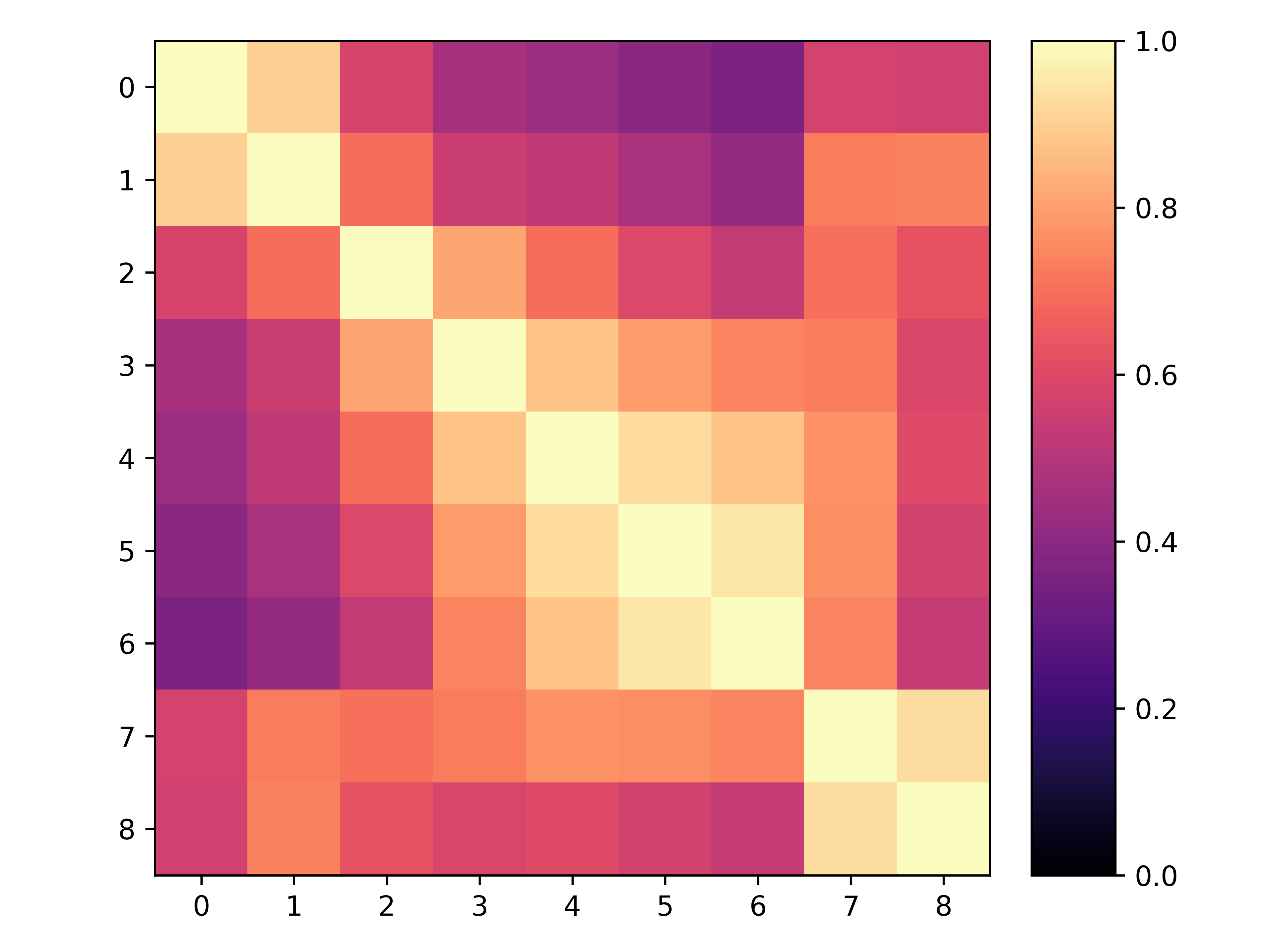}
        \vspace{5pt}
        \begin{picture}(0,0)
        \put(-125,6){\rotatebox{90}{{\footnotesize \texttt{Pre-trained Layers}}}}
        \put(-112,-7){{\footnotesize \texttt{Pre-trained Layers}}}
        \end{picture}
    \end{minipage}\hfill
    \begin{minipage}{.33\linewidth}
    \centering
        \includegraphics[width=0.95\textwidth]{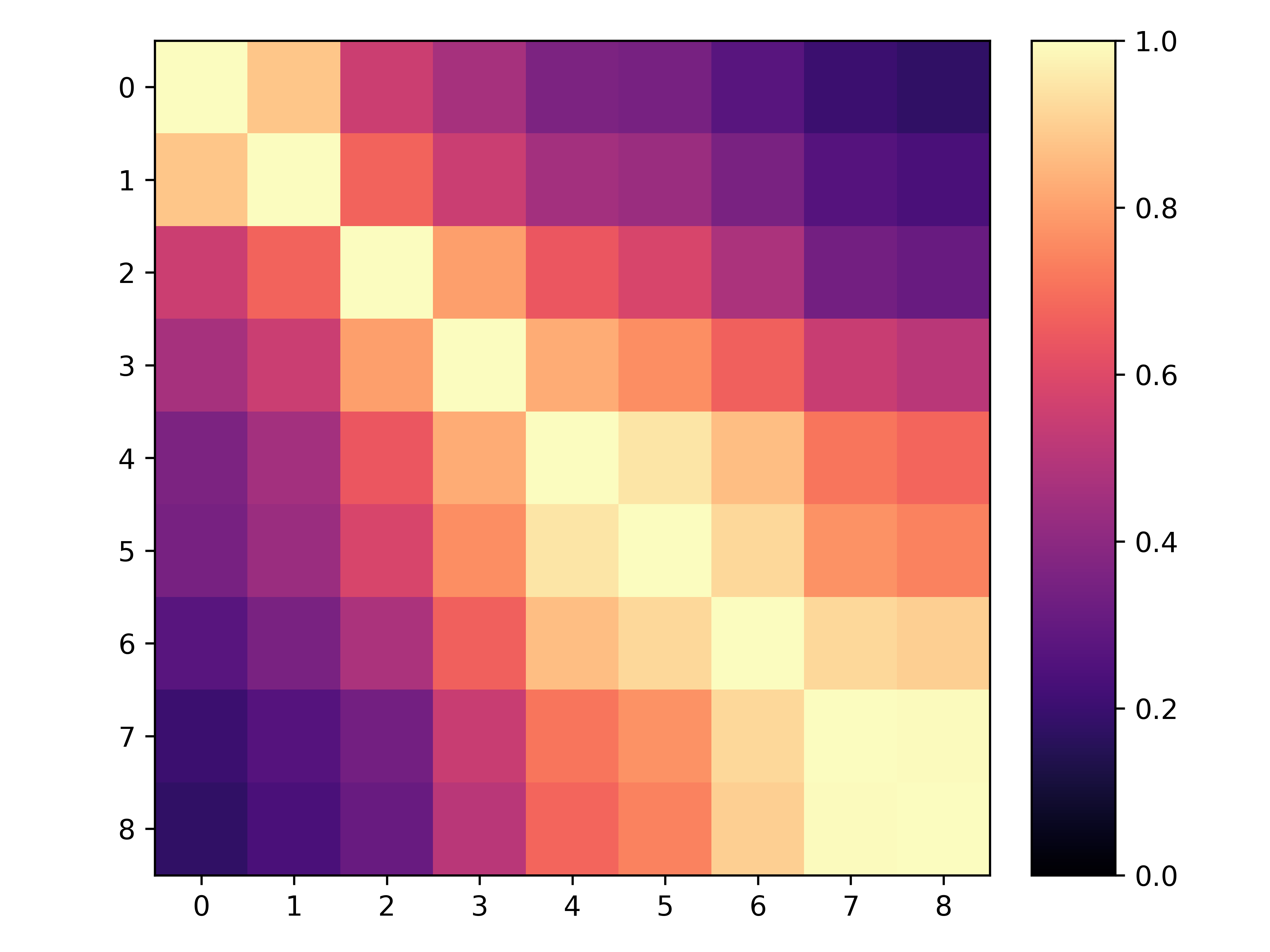}
        \vspace{5pt}
        \begin{picture}(0,0)
        \put(-125,6){\rotatebox{90}{{\footnotesize \texttt{Pre-trained Layers}}}}
        \put(-112,-7){{\footnotesize \texttt{Pre-trained Layers}}}
        \end{picture}
    \end{minipage}
  \caption{Layer-wise analysis of the three pre-training strategies, comparing the pre-trained representations as we delve deeper into the network (top row). MeSa effectively pre-trains the entire network, including the later layers. Left$\rightarrow$right: MP, MP+GP, MP+GP+SP (MeSa).}
  \label{fig:magis_cka_supp}
\end{figure}
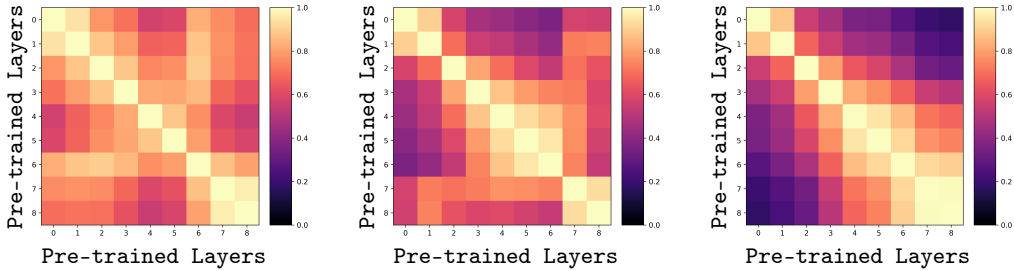

Here, we analyze the layer-wise representations of the MeSa pre-trained network to show that our pre-training strategy effectively pre-trains the entire network, including the later layers. To this end, we conduct a comparison between the layer 0 representation and the representations of layers 1-8, as illustrated in Figure~\ref{fig:magis_cka_supp} (top row). When utilizing masked pre-training (MP) alone, the later layers (5-8) do not provide significant benefits for downstream depth estimation since they become specialized for the reconstruction task, as evidenced by the increasing similarity values (lighter colors) past layer 4 in the top row. However, incorporating geometric pre-training in addition to masked pre-training (i.e., MP+GP) enables more effective representations to be learned deeper into the network, as indicated by the decreasing similarities (darker colors) up to layer 6. Nevertheless, layers 7-8 are still not fully utilized. By leveraging all three pre-training strategies (i.e., MP+GP+SP), we observe a monotonous decrease in similarities (darkening of colors) up to layer 8, indicating that the entire network learns distinct and high-level representations compared to layer 0. Hence, MeSa effectively pre-trains the entire network, including the later layers.
\section{Layer-wise Analysis of Different Pre-training Datasets}
In this section, we study the impact of different pre-training datasets on the layer-wise pre-trained features. For detailed configurations of the ImageNet and LSUN pre-trained networks, please refer to Section~\ref{sec:lsun_pretraining} in the main paper.

Similar to the analysis in Section~\ref{sec:sota_cka}, Figure~\ref{fig:imagenet_vs_lsun_p_p_cka} compares the representations of various layers within the pre-trained networks, whereas Figure~\ref{fig:imagenet_vs_lsun_p_ft_cka} compares the representations of the pre-trained networks and the (depth estimation) fine-tuned networks. The analysis in Figure~\ref{fig:imagenet_vs_lsun_p_p_cka} (top row) reveals that the intermediate layers of the LSUN pre-trained network exhibit lower similarities (darker colors) with layer 0 compared to the similarities of the corresponding layers of the ImageNet pre-trained network. Since earlier layers (e.g., layer 0) capture low-level features, this suggests that pre-training with LSUN learns more high-level representations in the middle layers that are beneficial for tasks requiring higher-level reasoning, as illustrated by the superior depth estimation results on NYUv2. This is further corroborated by Figure~\ref{fig:imagenet_vs_lsun_p_ft_cka}, which demonstrates that the LSUN pre-trained features are more effective for the downstream task than the ImageNet pre-trained features, as evidenced by the higher similarities (lighter colors) between the pre-trained and fine-tuned features along the diagonal line for LSUN compared to ImageNet. Since the images in LSUN have a smaller diversity compared to those in ImageNet, we hypothesize that pre-training on LSUN forces the network to reconstruct masked portions of similar images, thereby encouraging it to utilize more layers for higher level reasoning than for RGB reconstruction (Figure~\ref{fig:imagenet_vs_lsun_p_p_cka}).
\begin{figure}[!h]
\begin{minipage}{0.5\linewidth}
\centering
    \includegraphics[width=0.75\textwidth]{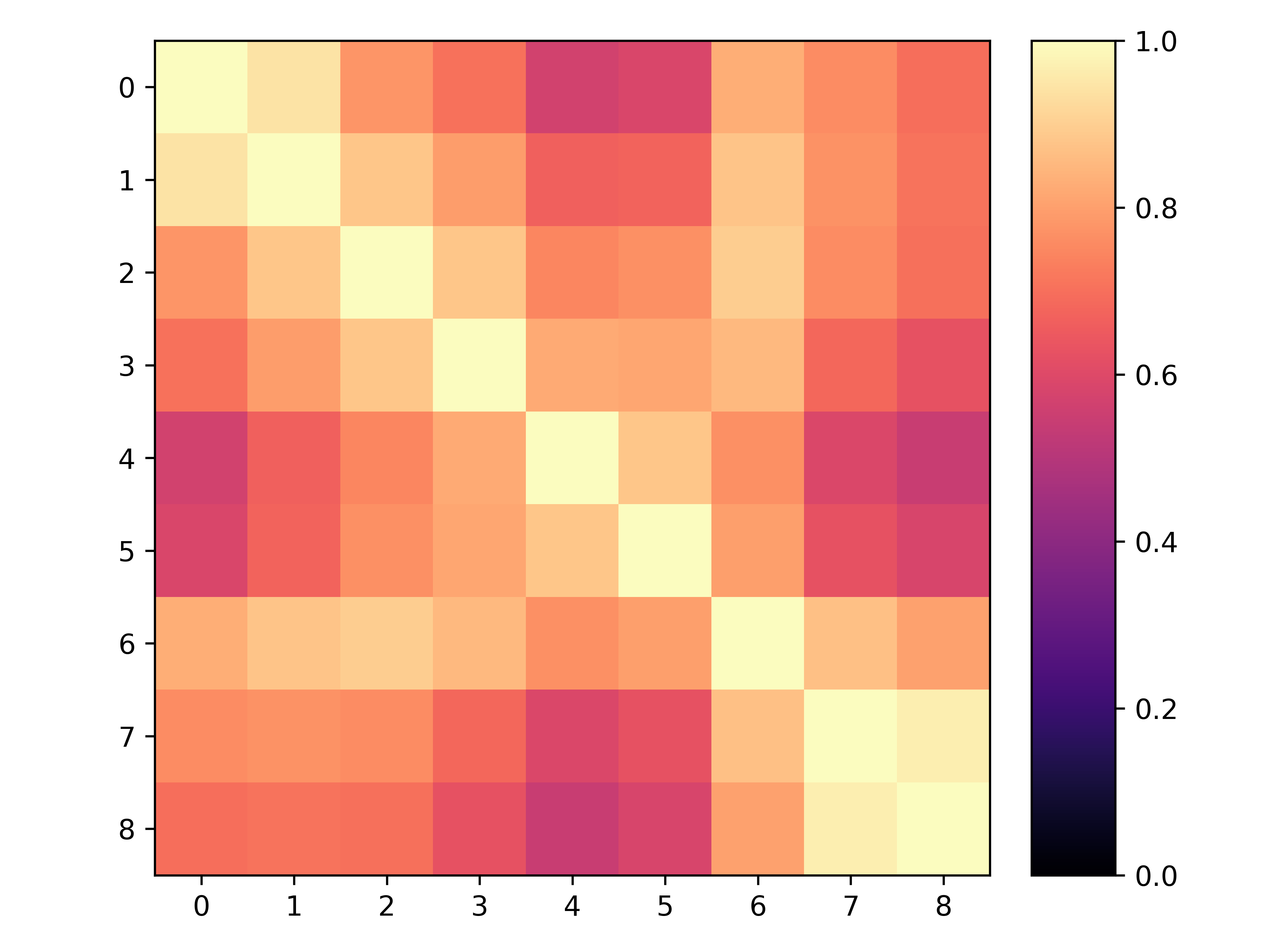}
    \vspace{10pt}
    \begin{picture}(0,0)
    \put(-150,15){\rotatebox{90}{{\footnotesize \texttt{Pre-trained Layers}}}}
    \put(-125,-8){{\footnotesize \texttt{Pre-trained Layers}}}
    \end{picture}
\end{minipage}\hfill
\begin{minipage}{0.5\linewidth}
\centering
    \includegraphics[width=0.75\textwidth]{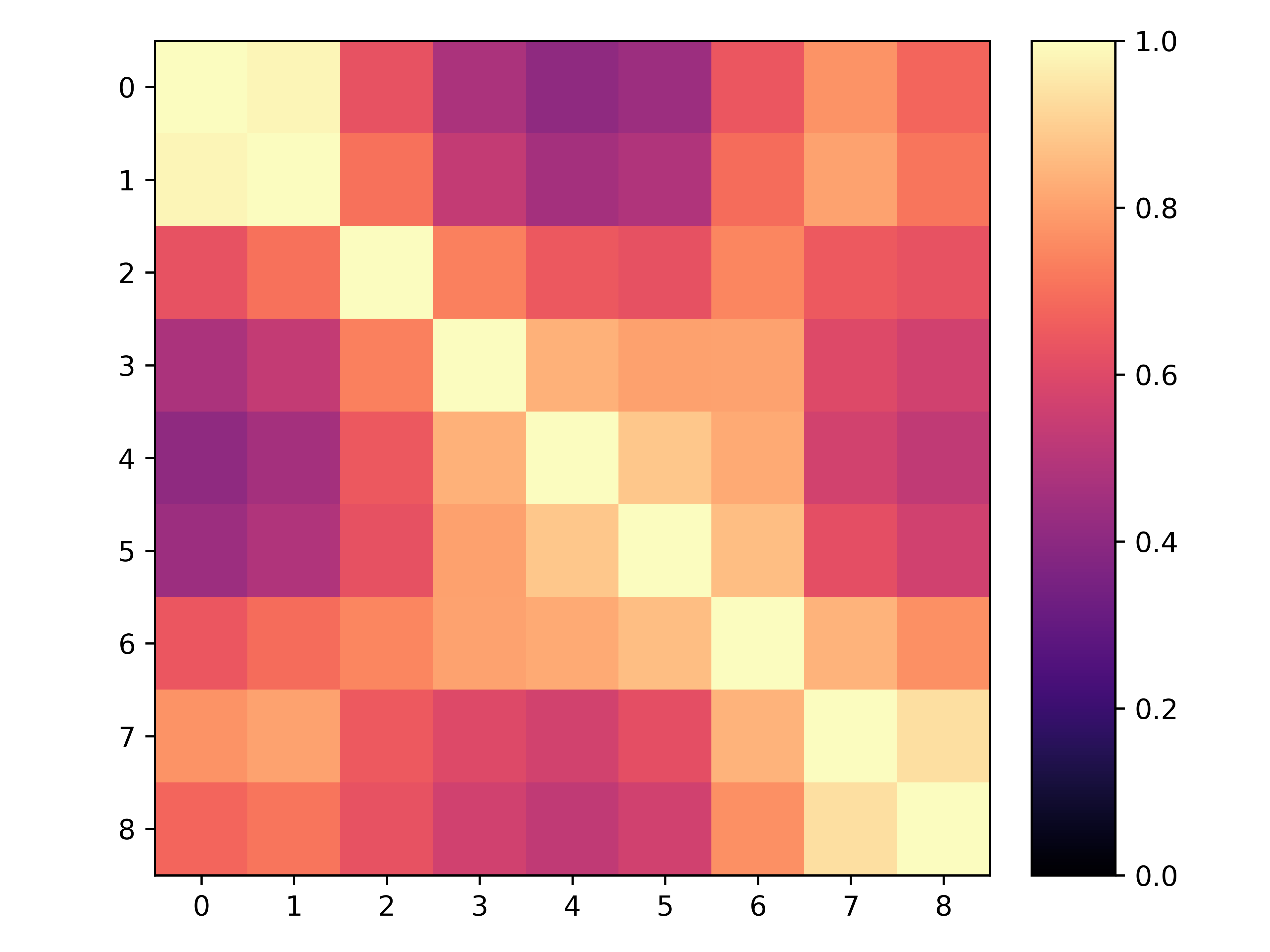}
    \vspace{10pt}
    \begin{picture}(0,0)
    \put(-150,15){\rotatebox{90}{{\footnotesize \texttt{Pre-trained Layers}}}}
    \put(-125,-8){{\footnotesize \texttt{Pre-trained Layers}}}
    \end{picture}
\end{minipage}
\caption{Layer-wise analysis comparing the representations of various layers within the pre-trained network. Left: ImageNet, right: LSUN. Pre-training with LSUN learns more effective higher-level features in the intermediate layers, as indicated by the lower similarities (darker colors) along the top row.
}
\label{fig:imagenet_vs_lsun_p_p_cka}
\end{figure}
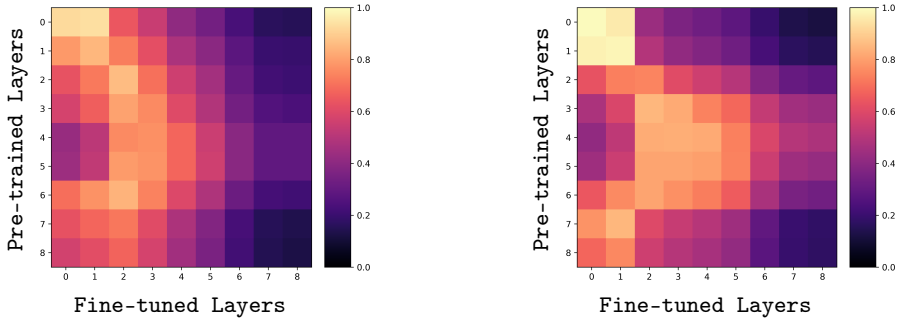
\begin{figure}[!h]
\begin{minipage}{0.5\linewidth}
\centering
    \includegraphics[width=0.75\textwidth]{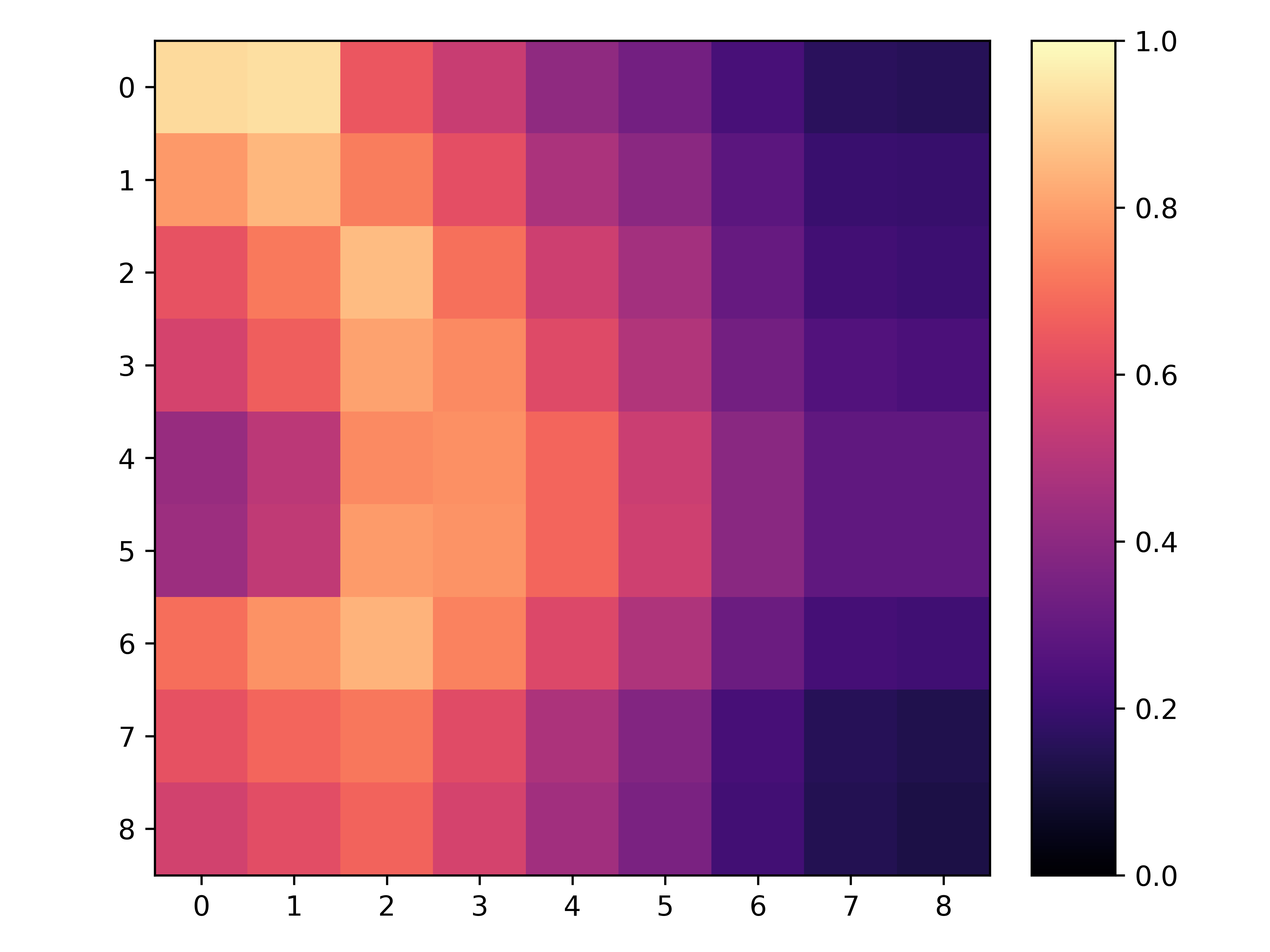}
    \vspace{10pt}
    \begin{picture}(0,0)
    \put(-150,15){\rotatebox{90}{{\footnotesize \texttt{Pre-trained Layers}}}}
    \put(-125,-8){{\footnotesize \texttt{Fine-tuned Layers}}}
    \end{picture}
\end{minipage}\hfill
\begin{minipage}{0.5\linewidth}
\centering
    \includegraphics[width=0.75\textwidth]{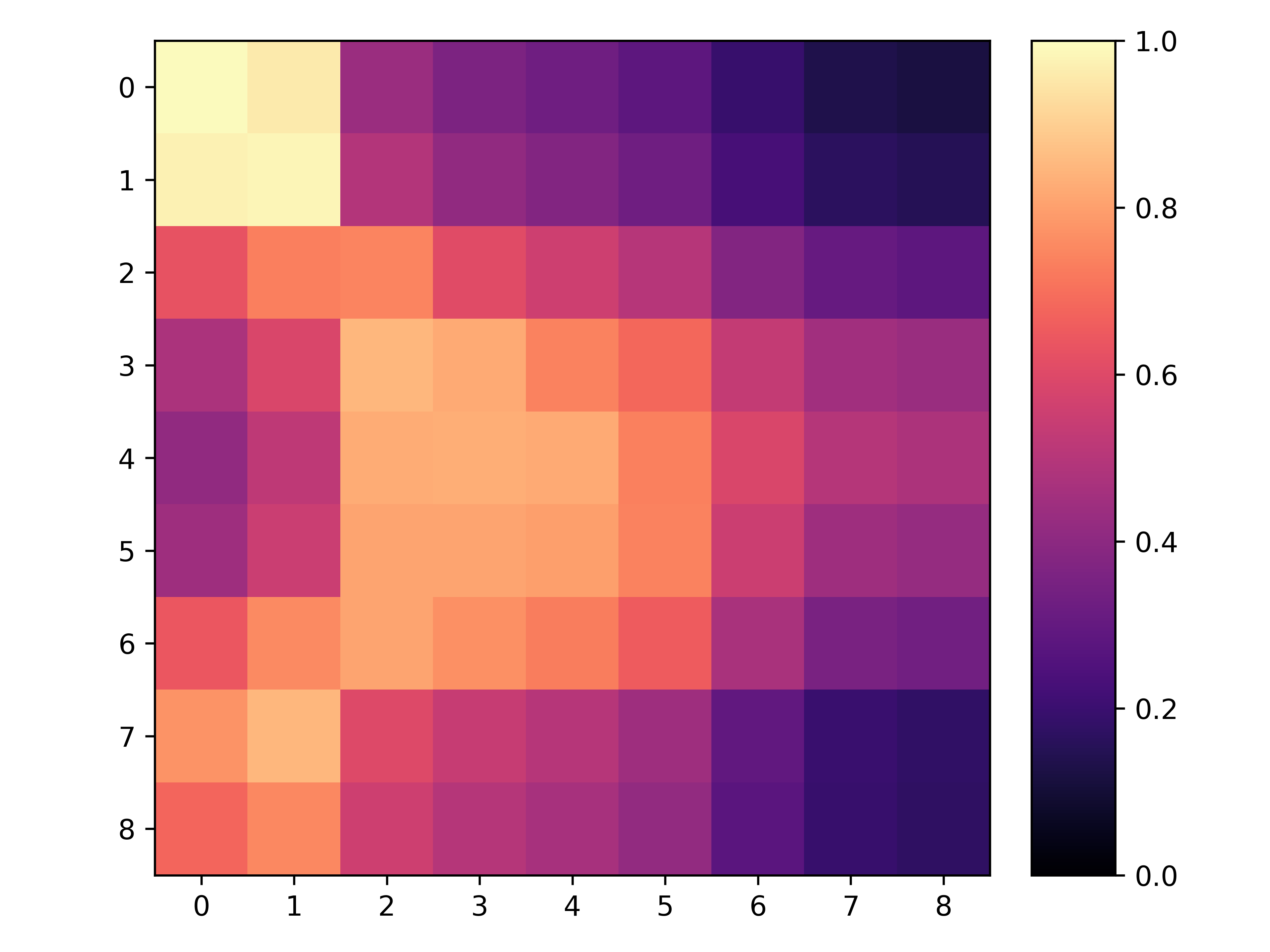}
    \vspace{10pt}
    \begin{picture}(0,0)
    \put(-150,15){\rotatebox{90}{{\footnotesize \texttt{Pre-trained Layers}}}}
    \put(-125,-8){{\footnotesize \texttt{Fine-tuned Layers}}}
    \end{picture}
\end{minipage}
\caption{Layer-wise analysis comparing the pre-trained features to the fine-tuned features. Left: ImageNet, right: LSUN. LSUN pre-trained features are more effective for the downstream task, as indicated by the higher similarities (lighter colors) between the pre-trained and fine-tuned features along the diagonal line.
}
\label{fig:imagenet_vs_lsun_p_ft_cka}
\end{figure}

\end{document}